\documentclass{article}

\PassOptionsToPackage{numbers, compress}{natbib}

     \usepackage[preprint]{neurips_data_2024}

\usepackage[utf8]{inputenc} 
\usepackage[T1]{fontenc}    
\usepackage{url}            
\usepackage{booktabs}       
\usepackage{amsfonts}       
\usepackage{nicefrac}       
\usepackage{microtype}      
\usepackage{xcolor}         
\usepackage{multirow}
\usepackage{multicol}
\definecolor{citeblue}{RGB}{48,111,186}
\usepackage[pagebackref=false,breaklinks=true,colorlinks=true,citecolor=citeblue,bookmarks=false]{hyperref}
\usepackage{enumitem}
\usepackage{amsmath}
\usepackage{cleveref}
\usepackage{comment}

\definecolor{nicergreen}{rgb}{0.13, 0.54, 0.13}
\newcommand\showdiff[1]{\textbf{\textcolor{nicergreen}{#1}}}
\usepackage{graphicx}
\usepackage{subcaption}
\usepackage{tikz}

\title{Revisiting Referring Expression Comprehension Evaluation in the Era of Large Multimodal Models}

\author{%
  Jierun Chen\thanks{Equal contribution.} \\
  HKUST \\
  \texttt{jcheneh@cse.ust.hk} \\
  \And
  Fangyun Wei\footnotemark[1]~~\thanks{Corresponding author.} \\
  Microsoft Research Asia\\
  \texttt{fawe@microsoft.com} \\
  \AND
  Jinjing Zhao\\
  The University of Sydney \\
  \texttt{jzha0100@uni.sydney.edu.au}\\
  \And
  Sizhe Song \\
  HKUST \\
  \texttt{ssongad@cse.ust.hk} \\
  \And
  Bohuai Wu \\
  HKUST \\
  \texttt{bwual@cse.ust.hk} \\
\And
  Zhuoxuan Peng \\
  HKUST \\
  \texttt{zpengac@cse.ust.hk} \\
\And
  S.-H. Gary Chan \\
  HKUST \\
  \texttt{gchan@cse.ust.hk} \\
\And
  Hongyang Zhang \\
  University of Waterloo \\
  \texttt{hongyang.zhang@uwaterloo.ca} \\
}

\begin{document}

\maketitle

\vspace{-4mm}
\begin{abstract}
\vspace{-3mm}
Referring expression comprehension (REC) involves localizing a target instance based on a textual description. Recent advancements in REC have been driven by large multimodal models (LMMs) like CogVLM, which achieved 92.44\% accuracy on RefCOCO. However, this study questions whether existing benchmarks such as RefCOCO, RefCOCO+, and RefCOCOg, capture LMMs' comprehensive capabilities. We begin with a manual examination of these benchmarks, revealing high labeling error rates: 14\% in RefCOCO, 24\% in RefCOCO+, and 5\% in RefCOCOg, which undermines the authenticity of evaluations. We address this by excluding problematic instances and reevaluating several LMMs capable of handling the REC task, showing significant accuracy improvements, thus highlighting the impact of benchmark noise. In response, we introduce Ref-L4, a comprehensive REC benchmark, specifically designed to evaluate modern REC models. Ref-L4 is distinguished by four key features: 1) a substantial sample size with 45,341 annotations; 2) a diverse range of object categories with 365 distinct types and varying instance scales from 30 to 3,767; 3) lengthy referring expressions averaging 24.2 words; and 4) an extensive vocabulary comprising 22,813 unique words. We evaluate a total of 24 large models on Ref-L4 and provide valuable insights. The cleaned versions of RefCOCO, RefCOCO+, and RefCOCOg, as well as our Ref-L4 benchmark and evaluation code, are available at \url{https://github.com/JierunChen/Ref-L4}.
\end{abstract}
\vspace{-4mm}
\section{Introduction}
\vspace{-3mm}
\label{sec:intro}



Referring expression comprehension (REC)~\cite{qiao2020referring,nagaraja2016modeling,zheng2022towards,kazemzadeh2014referitgame,pi2023perceptiongpt,zhang2019referring,wang2024towards} involves the task of localizing a specific target instance based on a given textual description. The advancement of REC has been significantly propelled by the superior language processing capabilities of large language models (LLMs)~\cite{touvron2023llama,touvron2023llama2,meta2024introducing,chiang2023vicuna,hao2022language,bai2023qwent,li2023textbooks,jiang2024mixtral,lewis2019bart}. This progress is particularly evident in the exceptional performance of large multimodal models (LMMs)~\cite{wang2023cogvlm,lin2023sphinx,gao2024sphinx,wang2023one,bai2023qwen,chen2023minigpt,wei2023lenna,he2024multi,wang2024git,huang2024relationvlm,zhao2024llm} on well-known benchmarks such as RefCOCO~\cite{yu2016modeling}, RefCOCO+~\cite{yu2016modeling}, and RefCOCOg~\cite{mao2016generation}. These models have demonstrated remarkable accuracy, with CogVLM~\cite{wang2023cogvlm}, for instance, achieving an impressive accuracy rate of 92.44\% on the RefCOCO benchmark.

\vspace{-0.5mm}
This paper begins with a critical question: do existing REC benchmarks truly capture the comprehensive capabilities of LMMs? The foundational benchmarks, RefCOCO~\cite{yu2016modeling}, RefCOCO+~\cite{yu2016modeling}, and RefCOCOg~\cite{mao2016generation}, were introduced sequentially in 2015, 2016, and 2016, respectively. In RefCOCO, the referring expressions are notably succinct, ranging from single words like ``\textit{lady}'' and ``\textit{yellow}'' to brief descriptions such as ``\textit{far left person}'' and ``\textit{white shirt}''. RefCOCO+ intentionally excludes locational prepositions commonly found in RefCOCO, favoring short yet semantically rich expressions like ``\textit{plastic cup with just ice}'' and ``\textit{man on screen}''. Conversely, RefCOCOg provides more elaborate annotations, including examples such as ``\textit{a table of food, with plates, a pizza, pitchers, and glasses}'' and ``\textit{a red and white checkered table with two wooden chairs}''. These variations highlight the evolution and complexity of referring expressions across different benchmarks, raising the question of whether they can effectively assess the nuanced capabilities of modern LMMs in understanding diverse linguistic inputs and associating languages with visual elements.

\vspace{-1mm}
\textbf{Labeling Error Rates of Existing Benchmarks.} To begin, we manually assess the labeling error rates of the validation and test sets in RefCOCO, RefCOCO+, and RefCOCOg, discovering a high error rate across these benchmarks. The labeling errors include, typos, misalignment between referring expressions and target instances, as well as inaccurate bounding box annotations, as depicted in Section~\ref{sec:label_error}. As illustrated in Table~\ref{tab:refcoco_error}, the labeling error rates for RefCOCO, RefCOCO+, and RefCOCOg are 14\%, 24\%, and 5\%, respectively, indicating that evaluations performed on these benchmarks may lack authenticity.

\begin{table}[!t]
\centering
\small 
\caption{Statistics of the labeling error rates for RefCOCO, RefCOCO+, and RefCOCOg, respectively. For each benchmark, the statistics are conducted on the combination of the validation and test sets.}
\label{tab:refcoco_error}
\begin{tabular}{l|ccc}
\toprule
Benchmark  & \multicolumn{1}{c}{Annotations} & \multicolumn{1}{c}{Errors} & \multicolumn{1}{c}{Labeling Error Rate} \\ \midrule
RefCOCO~\cite{yu2016modeling}   & 21,586                   & 3,054                          & 14\%                           \\
RefCOCO+~\cite{yu2016modeling}  & 21,373                   & 5,201                          & 24\%                           \\
RefCOCOg~\cite{mao2016generation}  & 14,498                   & 675                            & 5\%                           \\ \bottomrule
\end{tabular}
\vspace{-1mm}
\end{table}

\begin{table}[!t]
\centering
\small
\setlength{\tabcolsep}{5pt} 
\vspace{-4mm}
\caption{
The performance of four LMMs capable of handling the REC task on both the cleaned and original versions of the RefCOCO, RefCOCO+, and RefCOCOg benchmarks, using the conventional accuracy as the evaluation metric. The evaluation is performed on the combination of the validation and test sets for each benchmark. $\dagger$: models fine-tuned on the specific dataset.
}
\label{tab:reevaluate}
\begin{tabular}{l|lllll}
\toprule
Benchmark
& \begin{tabular}[c]{@{}l@{}}ONE-\\PEACE$^\dagger$~\cite{wang2023one}\end{tabular}
& OFA-L$^\dagger$~\cite{wang2022ofa}
& OFA-L~\cite{wang2022ofa}
& Qwen-VL~\cite{bai2023qwen}
& \begin{tabular}[c]{@{}l@{}}CogVLM-\\ Grounding~\cite{wang2023cogvlm}\end{tabular} \\ \midrule
RefCOCO~\cite{yu2016modeling}            & 92.15         & 89.85         & 85.13         & 88.51         & 92.44            \\
RefCOCO (Cleaned)  & 94.11 \showdiff{(+1.96)} & 92.06 \showdiff{(+2.22)} & 87.95 \showdiff{(+2.81)} & 90.68 \showdiff{(+2.18)} & 94.58 \showdiff{(+2.13)}    \\ \midrule
RefCOCO+~\cite{yu2016modeling}           & 88.14         & 85.06         & 77.56         & 82.52         & 88.55            \\
RefCOCO+ (Cleaned) & 90.79 \showdiff{(+2.66)} & 87.38 \showdiff{(+2.32)} & 80.50 \showdiff{(+2.94)} & 85.60 \showdiff{(+3.08)} & 91.43 \showdiff{(+2.87)}    \\ \midrule
RefCOCOg~\cite{mao2016generation}           & 89.18         & 84.77         & 79.25         & 85.11         & 90.67            \\
RefCOCOg (Cleaned) & 90.75 \showdiff{(+1.57)} & 86.39 \showdiff{(+1.62)} & 80.89 \showdiff{(+1.64)} & 86.79 \showdiff{(+1.68)} & 92.36 \showdiff{(+1.68)}    \\ \bottomrule
\end{tabular}
\vspace{-3mm}
\end{table}

\vspace{-1mm}
\textbf{Reevaluation on RefCOCO, RefCOCO+ and RefCOCOg.} In response, we manually exclude the problematic instances from the validation and test sets of RefCOCO, RefCOCO+, and RefCOCOg. Subsequently, we reevaluate four LMMs capable of handling the REC task—namely ONE-PEACE~\cite{wang2023one}, OFA-L~\cite{wang2022ofa}, Qwen-VL~\cite{bai2023qwen}, and CogVLM-Grounding~\cite{wang2023cogvlm}—on both the cleaned and original versions of these datasets, as shown in Table~\ref{tab:reevaluate}. Across all models and cleaned benchmarks, we observe a significant accuracy improvement, ranging from 1.57 to 3.08, compared to their performance on the original versions. This demonstrates that noise in the benchmarks has impacted the models' true capabilities. \textit{To support further research in the REC field, we release the cleaned versions of RefCOCO, RefCOCO+, and RefCOCOg.}

\begin{table}[!t]
\centering
\small
\setlength{\tabcolsep}{3.5pt} 
\caption{Comparison between our Ref-L4 benchmark and other REC benchmarks, including RefCOCO~\cite{yu2016modeling}, RefCOCO+~\cite{yu2016modeling}, and RefCOCOg~\cite{mao2016generation}. For the latter three benchmarks, we combine their validation and test sets for statistics. The instance size and image size are represented by their respective square roots. Avg. length: average length of annotations. Vocab.: vocabulary size.}
\label{tab:dataset_comparison}
\begin{tabular}{l|cccccccc}
\toprule
\multirow{2}{*}{Benchmark} &
  
  \multicolumn{1}{c}{\multirow{2}{*}{Images}} &
  \multicolumn{1}{c}{\multirow{2}{*}{Instances}} &
  \multicolumn{1}{c}{\multirow{2}{*}{Annotations}} &
  \multicolumn{1}{c}{\multirow{2}{*}{Categories}} &
  Avg. &
  Instance &
  Image &
  \multicolumn{1}{c}{\multirow{2}{*}{Vocab.}} \\ 
 &
  \multicolumn{1}{l}{} &
  \multicolumn{1}{l}{} &
  \multicolumn{1}{l}{} &
  \multicolumn{1}{l}{} &
  Length &
  Size &
  Size & \\ 
  \midrule
RefCOCO~\cite{yu2016modeling}       & 3,000 & 7,596  & 21,586 & 71  & 3.6  & 105~-~607 & 230~-~640  & 3,525   \\
RefCOCO+~\cite{yu2016modeling}    & 3,000 & 7,578  & 21,373 & 71  & 3.6  & 105~-~607 & 230~-~640  & 4,387   \\
RefCOCOg~\cite{mao2016generation}          & 3,900 & 7,596  & 14,498 & 78   & 8.4  & 83~-~610 & 277~-~640 & 5,050   \\ \midrule
Ref-L4 (Ours)       & 9,735 & 18,653 & 45,341 & 365 & 24.2 & 30~-~3,767 & 230~-~6,606 & 22,813 \\ \bottomrule
\end{tabular}
\end{table}

\begin{figure}[!t]
\centering
\includegraphics[width=\textwidth]{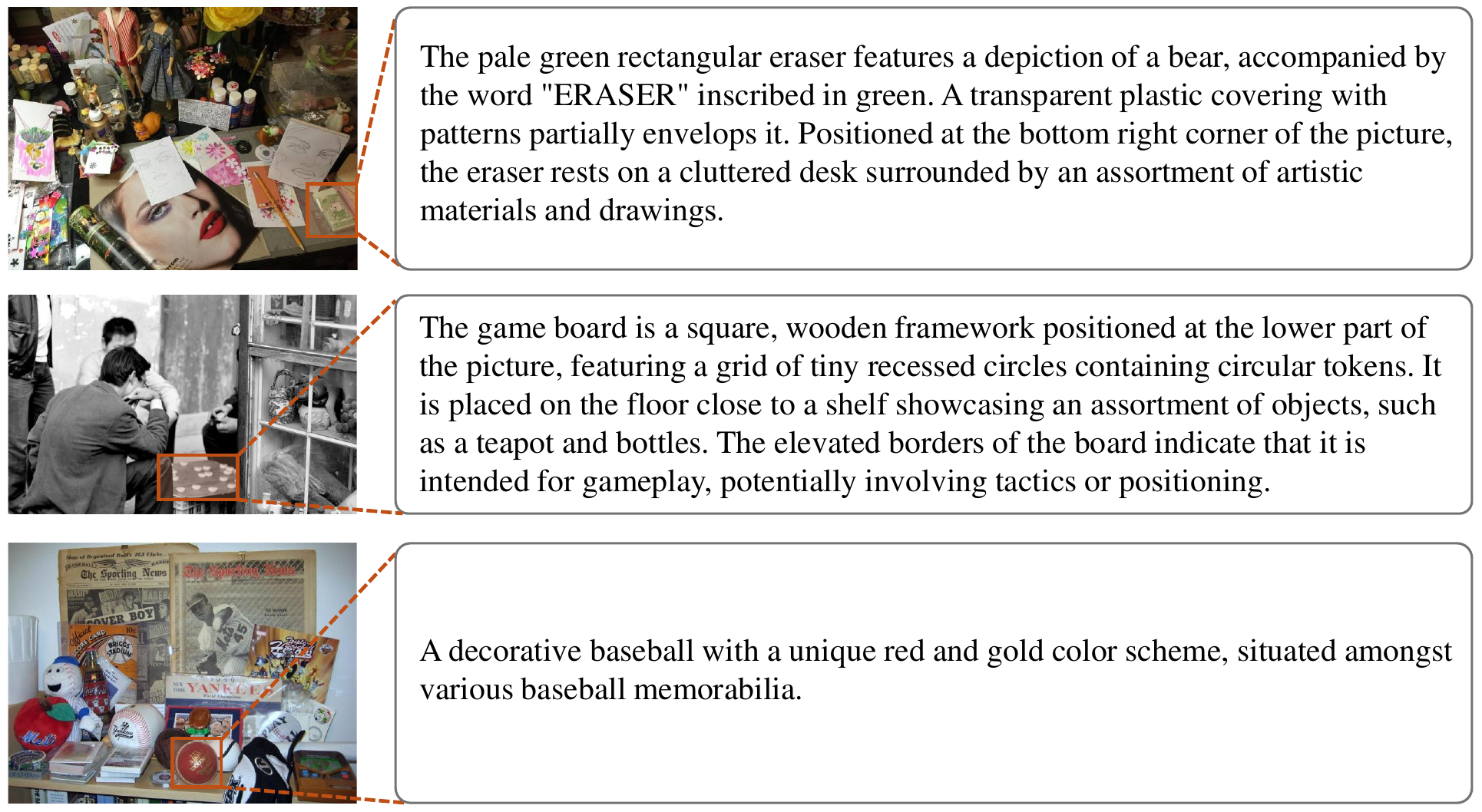}
\vspace{-4mm}
\caption{Examples from our Ref-L4 benchmark. We offer a detailed referring expression for each target instance represented by a bounding box. Zoom in for better visualization.}
\label{fig:examples}
\end{figure}

\vspace{-1mm}
\textbf{Ref-L4: A Comprehensive REC Benchmark for Modern LMM Evaluation.} We present Ref-L4, where L4 signifies four key aspects: a \underline{L}arge number of testing samples, \underline{L}arge diversity in object categories and instance scales, \underline{L}ong referring expressions, and a \underline{L}arge vocabulary. These features make Ref-L4 a comprehensive benchmark for assessing the REC capabilities of contemporary LMMs. Table~\ref{tab:dataset_comparison} provides a detailed comparison between Ref-L4 and other benchmarks including RefCOCO, RefCOCO+, and RefCOCOg. Our Ref-L4 benchmark stands out due to the following characteristics:
\vspace{-3mm}
\begin{itemize}
\vspace{-3mm}
    \item \textit{Large-Scale.} Ref-L4 includes $9,735$ images, $18,653$ unique instances, and a total of $45,341$ annotations, significantly surpassing RefCOCO, RefCOCO+, and RefCOCOg. For instance, RefCOCOg offers $3,900$ images, $7,596$ instances, and $14,498$ annotations. 
    \vspace{-1mm}
    \item  \textit{High Diversity.} Ref-L4 features $365$ unique categories. Since the RefCOCO series derive from the COCO 2014 dataset, they encompass up to $78$ categories. Additionally, our benchmark covers a wider range of instance scales, from $30$ to $3,767$, measured by the square root of the instance area.
    \vspace{-1mm}
    \item \textit{Lengthy Referring Expressions.} Each referring expression in Ref-L4 is a detailed description of a specific instance, with lengths ranging from $3$ to $117$ words and an average of $24.2$ words. In comparison, the average annotation lengths in RefCOCO, RefCOCO+, and RefCOCOg are $3.6$, $3.6$, and $8.4$ words, respectively. Examples can be found in Figure~\ref{fig:examples}.
    \item \textit{Extensive Vocabulary.} Due to the detailed nature of the referring expressions, Ref-L4 boasts a large vocabulary of $22,813$ words, which is four to six times larger than those of RefCOCO, RefCOCO+, and RefCOCOg.
\end{itemize}

\textbf{Evaluation on Ref-L4.} We conduct an evaluation of 24 representative LMMs that can perform the REC task. In addition to the standard accuracy metric, which considers predictions with an IoU greater than 0.5 as accurate (Acc$_{0.5}$), we also report accuracies at higher IoU thresholds: Acc$_{0.75}$ and Acc$_{0.9}$. Furthermore, we introduce a mean accuracy (mAcc), calculated as the average accuracy from Acc$_{0.5}$ to Acc$_{0.9}$ in increments of 0.05. To gain deeper insights into the models' capabilities, we conduct a detailed analysis of REC performance across different instance scales and categories. \textit{The Ref-L4 benchmark and the evaluation code are available at \url{https://github.com/JierunChen/Ref-L4}.}

\vspace{-2mm}
\section{Related Work}
\vspace{-2mm}
\label{sec:related}

\textbf{REC and Its Benchmarks.} Referring Expression Comprehension (REC)~\cite{qiao2020referring,nagaraja2016modeling,zheng2022towards,kazemzadeh2014referitgame,pi2023perceptiongpt,zhang2019referring} is a task that involves identifying a specific object within an image based on a given referring expression. Unlike object detection~\cite{lin2014microsoft,krishna2017visual,shao2019objects365,redmon2016you,carion2020end}, which operates within fixed categories and a single visual modality, REC necessitates understanding free-form text to locate objects of any category. Phrase Grounding~\cite{plummer2015flickr30k,wu2020phrasecut,gupta2019lvis,liu2023grounding,li2022grounded,zhang2022glipv2,wang2020maf} is similar but typically involves shorter phrases and identifies multiple regions, whereas REC requires parsing longer expressions to pinpoint a single unique region. This complexity makes REC an ideal task for evaluating emerging large multimodal models. Current REC benchmarks such as RefCOCO~\cite{yu2016modeling}, RefCOCO+\cite{yu2016modeling}, and RefCOCOg\cite{mao2016generation} include tens of thousands of annotations but are limited by their short expression lengths—averaging 3.6, 3.6, and 8.4 words, respectively. Additionally, they encompass fewer than 80 categories, lacking real-world diversity. Other REC benchmarks ~\cite{liu2019clevr,chen2023advancing,qiu2022refcrowd,chen2020cops,wang2024beyond,kurita2023refego,wang2020give,cirik2020refer360,bu2022scene,gao2023textrec,de2017guesswhat,jia2024sceneverse} are often designed for specific scenarios. For example, CLEVR-Ref+\cite{liu2019clevr} focuses on simple objects like boxes, spheres, and cylinders. SK-VG\cite{chen2023advancing} integrates prior scene knowledge as additional input, while RefCrowd~\cite{qiu2022refcrowd} targets identifying a person within a crowd. By contrast, we introduce Ref-L4, a more general and comprehensive benchmark encompassing 365 categories and 45,341 annotations. Ref-L4 features expressions averaging 24.2 words and a vocabulary of 22,813 words, facilitating the accurate evaluation of REC models on complex expressions and diverse objects.

\textbf{REC Models.} The evolution of REC models has transitioned from specialized models~\cite{kamath2021mdetr,yu2018mattnet,liu2017referring,su2019vl,zheng2019reasoning,yan2023universal,zou2024segment} to generalist models or large multimodal models (LMMs)\cite{wang2023cogvlm,lin2023sphinx,gao2024sphinx,wang2023one,bai2023qwen,chen2023minigpt,wei2023lenna,zhang2024ferret,zhan2023griffon,zhan2024griffon,pramanick2023jack,zhang2023llava,wang2024visionllm,shen2024groundvlp,ma2024groma,qi2024generalizable,kosarevapushing}. Notable examples of these LMMs include CogVLM-Grounding\cite{wang2023cogvlm}, SPHINX~\cite{lin2023sphinx,gao2024sphinx}, ONE-PEACE~\cite{wang2023one}, Qwen-VL-Chat~\cite{bai2023qwen}, MiniGPTv2~\cite{chen2023minigpt}, and Lenna~\cite{wei2023lenna}. These models, benefiting from larger model sizes and extensive training on diverse datasets, exhibit remarkable performance on conventional REC datasets. For example, CogVLM-Grounding achieves an accuracy of $94.58\%$ on RefCOCO (cleaned). Additionally, the performance gap among models is shrinking, with many LMMs surpassing $90\%$ accuracy. This performance saturation raises concerns about the adequacy of current REC benchmarks for making meaningful comparisons. In response, we propose Ref-L4, a more comprehensive and challenging benchmark. We have also conducted rigorous evaluations of 24 LMM models, offering holistic comparisons that highlight their weaknesses and suggest directions for improvement.

\vspace{-2mm}
\section{Ref-L4}
\vspace{-2mm}
\label{sec:method}

\begin{figure}[!t]
\centering
\includegraphics[width=\textwidth]{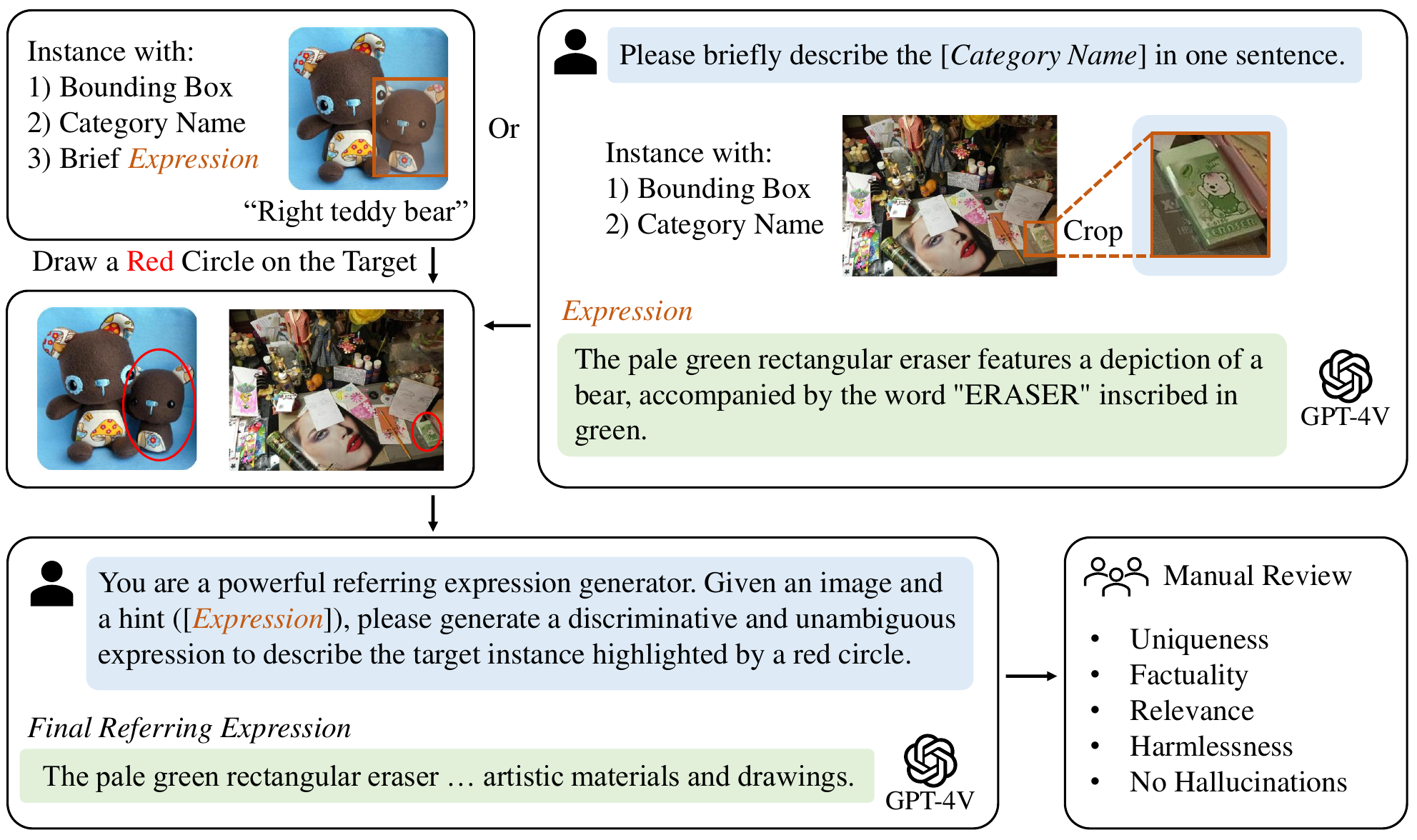}
\vspace{-4mm}
\caption{Pipeline of generating a referring expression for a target instance.}
\label{fig:overview}
\end{figure}

\subsection{Benchmark Creation}
\vspace{-2mm}
\textbf{Data Sources.} 
Our benchmark is derived from two sources: 1) our cleaned validation and test sets of the RefCOCO~\cite{yu2016modeling}, RefCOCO+~\cite{yu2016modeling}, and RefCOCOg~\cite{mao2016generation} datasets; and 2) the test set from the large-scale object detection dataset Objects365~\cite{shao2019objects365}. The Objects365 dataset provides a broader range of categories, varying instance sizes, higher image resolutions, and more intricate scenes. In the RefCOCO series, each instance includes a bounding box, a category name, and an extremely brief expression like ``right teddy bea''. In contrast, the Objects365 benchmark labels each instance with mainly a bounding box and the relevant category.

For the RefCOCO (cleaned) series, we begin by consolidating duplicate images and instances, resulting in a subset of $6,502$ images containing $14,186$ unique instances. For Objects365, we select samples from its testing set based on several criteria: 1) each image has both height and width greater than 800 pixels; 2) each image is sufficiently complex, containing more than 10 categories and 20 instances; 3) each instance has a square normalized size $\sqrt{(hw)/(HW)}$ greater than 0.05, where $(h,w)$ represents the instance size and $(H,W)$ denotes the image size; 4) we randomly sample $N$ instances for each of the 365 classes defined in Objects365, with $N=\min(35, \text{the number of instances for the specific class})$; 5) we review and exclude instances with erroneous bounding box annotations or those difficult to describe uniquely. For a few rare classes, we relax criterion-1 to 512 pixels and criterion-2 to 10 instances. Consequently, we collect $3,233$ images and $4,467$ instances from Objects365. Overall, our Ref-L4 benchmark comprises $9,735$ images and $18,653$ instances, sourced from the RefCOCO series and Objects365.

\textbf{Referring Expression Generation.} Given a target instance and its corresponding image, we leverage GPT-4V with human reviewers in the loop to generate its precise and detailed referring expressions. Figure~\ref{fig:overview} illustrates the three-step generation process:

\textit{Step-1}: Each instance in the Objects365 dataset is linked to a bounding box and a \textit{category name}. We begin by cropping these instances from the original images. Next, we input each cropped area along with the prompt detailed in Section~\ref{sec:prompt_instance-level} into GPT-4V to produce a context-independent description. For instances from the RefCOCO series, this step is omitted as each instance already has a brief expression.

\textit{Step-2}: Drawing inspiration from recent studies on GPT-4V~\cite{yang2023dawn}, where GPT-4V is able to pay more attention to instances highlighted by a red circle within an image, we similarly encircle the target instance in red to facilitate GPT-4V in generating a context-aware referring expression. Following this, as depicted in Figure~\ref{fig:overview}, we process the image and use the prompt outlined in Section~\ref{sec:prompt_contextual} to generate a context-aware referring expression for each instance. We instruct GPT-4V to describe various features such as color, size, position, and context. Additionally, we provide a hint (the context-independent description from Step-1) in the prompt to mitigate hallucination issues, resulting in more accurate descriptions.

\textit{Step-3}: We manually review all generated referring expressions to correct any hallucination issues. We ensure that each expression uniquely describes the instance and is factual, accurate, and harmless.

\begin{figure}[!t]
    \centering
        \begin{subfigure}[b]{0.49\textwidth}
        \centering
        \includegraphics[width=\textwidth]{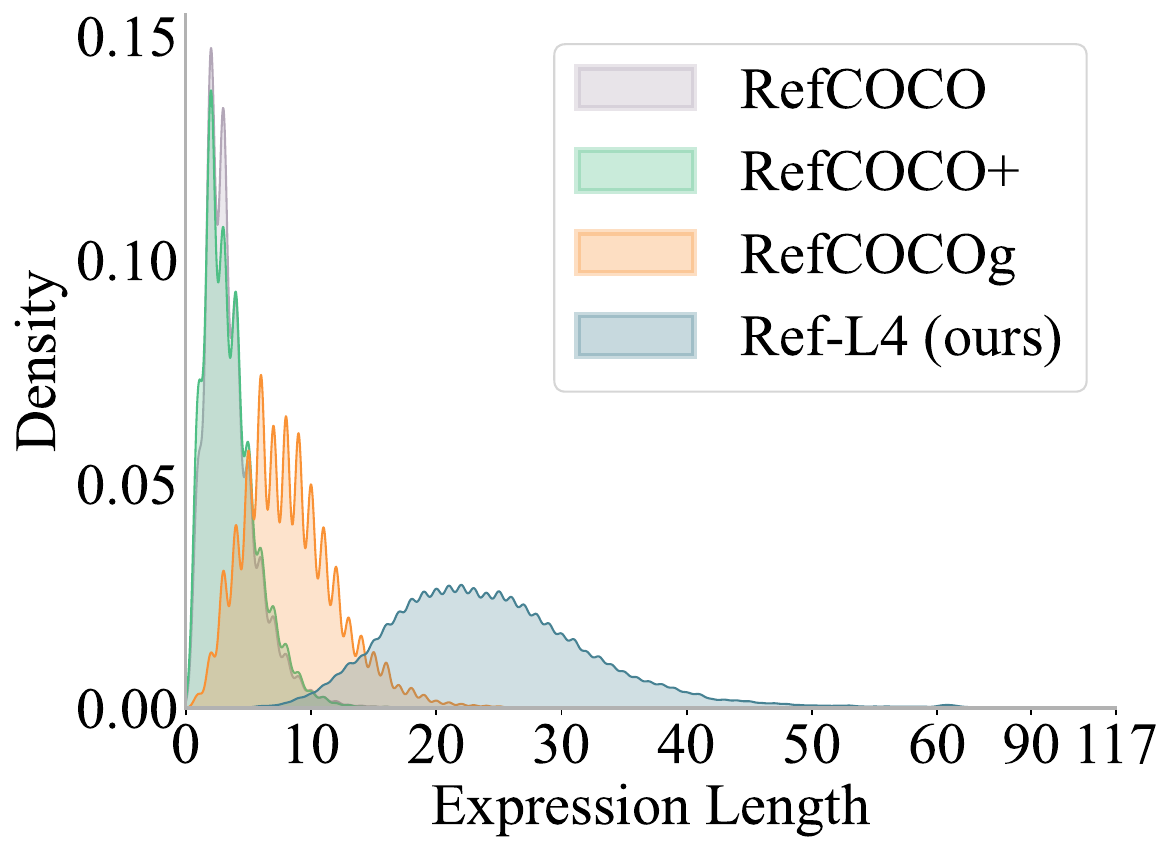}
        \caption{The distribution of expression length.}
        \label{fig:exp_length_distri}
    \end{subfigure}
    \begin{subfigure}[b]{0.49\textwidth}
        \centering
        \includegraphics[width=\textwidth]{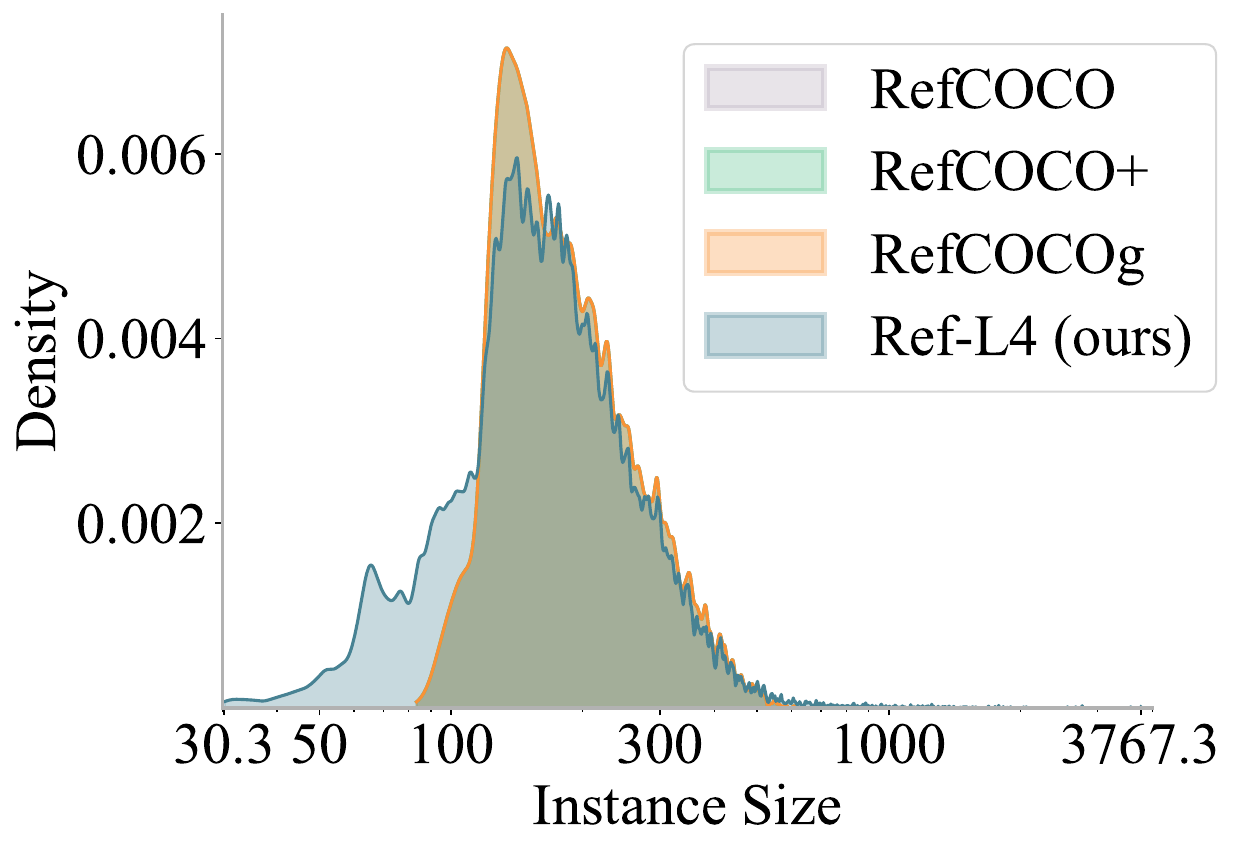}
        \caption{The distribution of instance size.}
        \label{fig:box_size_distri}
    \end{subfigure}
    \begin{subfigure}[t]{0.98\textwidth}
        \centering
        \includegraphics[width=\textwidth]{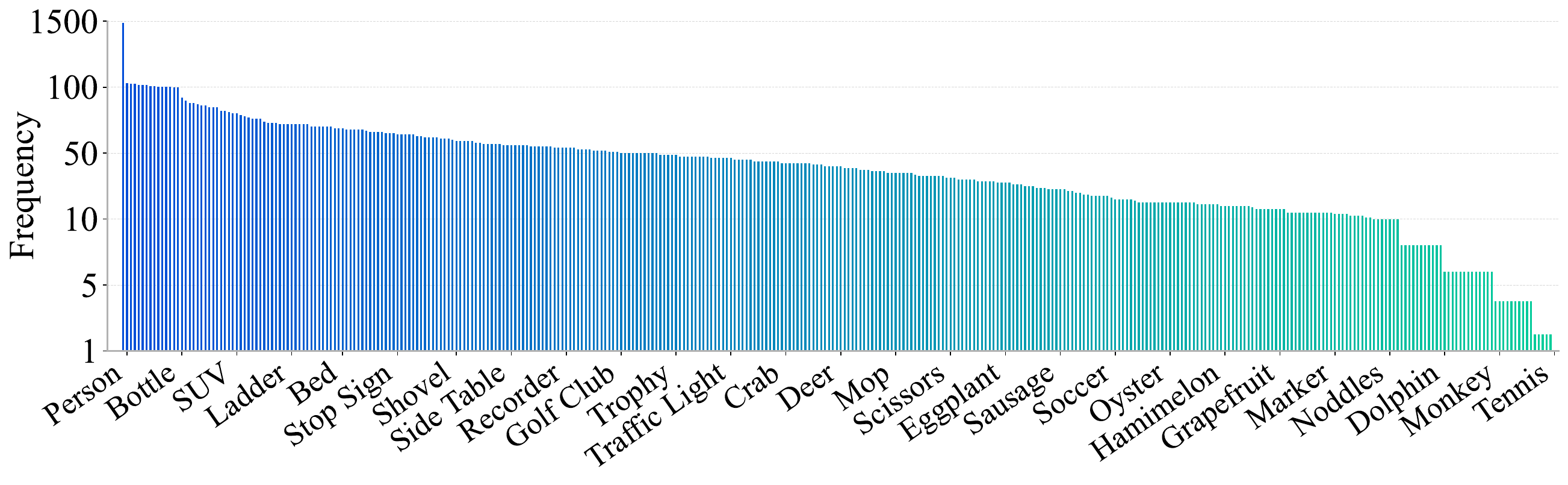}
        \vspace{-4mm}
        \caption{The distribution of instance numbers over 365 categories.}
        \label{fig:category_distribution}
    \end{subfigure}
    \caption{Analysis of referring expression length, instance size, and category distribution.}
    \label{fig:analysis}
\end{figure}

\begin{figure}[!t]
    \centering
    \begin{subfigure}[b]{0.25\textwidth}
        \centering
        \includegraphics[width=\textwidth]{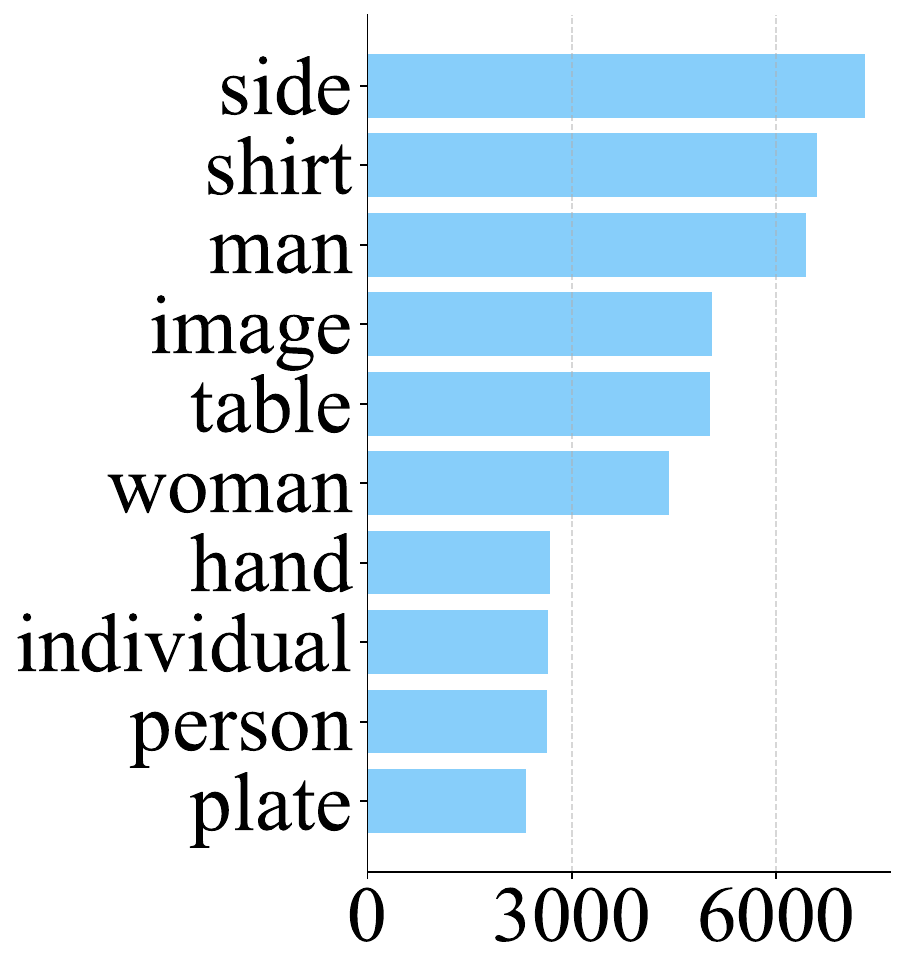}
        \caption{Nouns.}
    \end{subfigure}
    \begin{subfigure}[b]{0.25\textwidth}
        \centering
        \includegraphics[width=\textwidth]{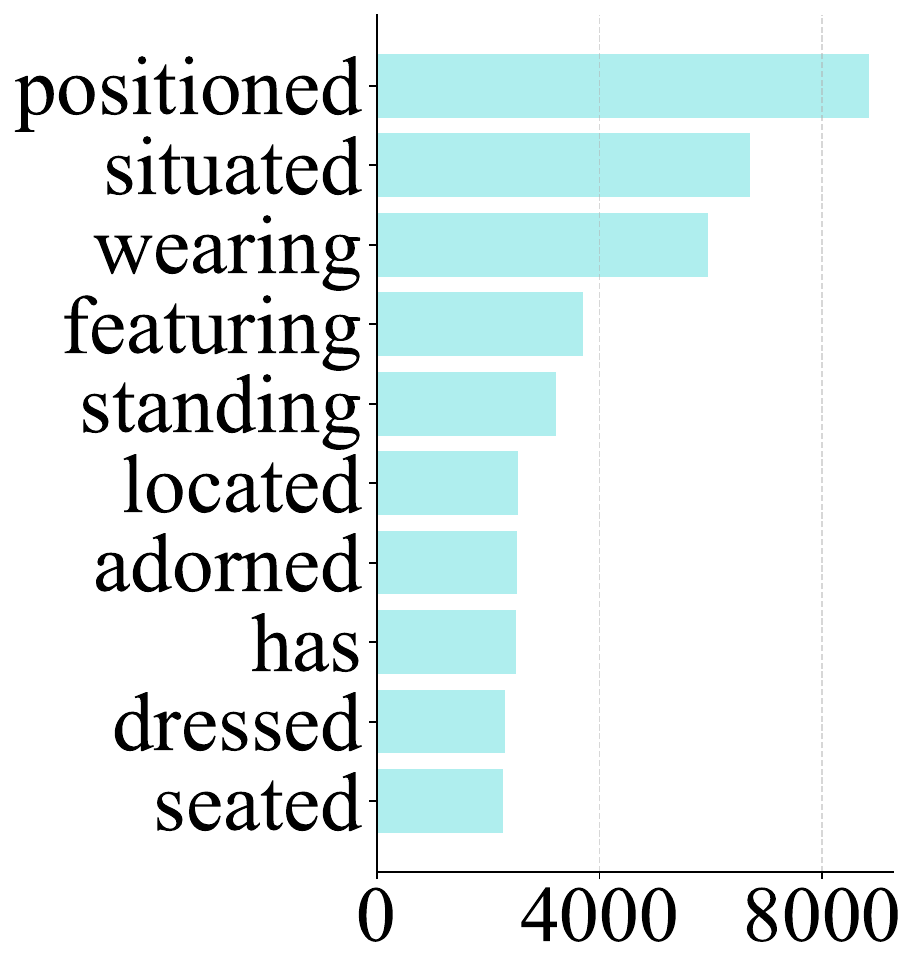}
        \caption{Verbs.}
    \end{subfigure}
    \begin{subfigure}[b]{0.25\textwidth}
        \centering
        \includegraphics[width=\textwidth]{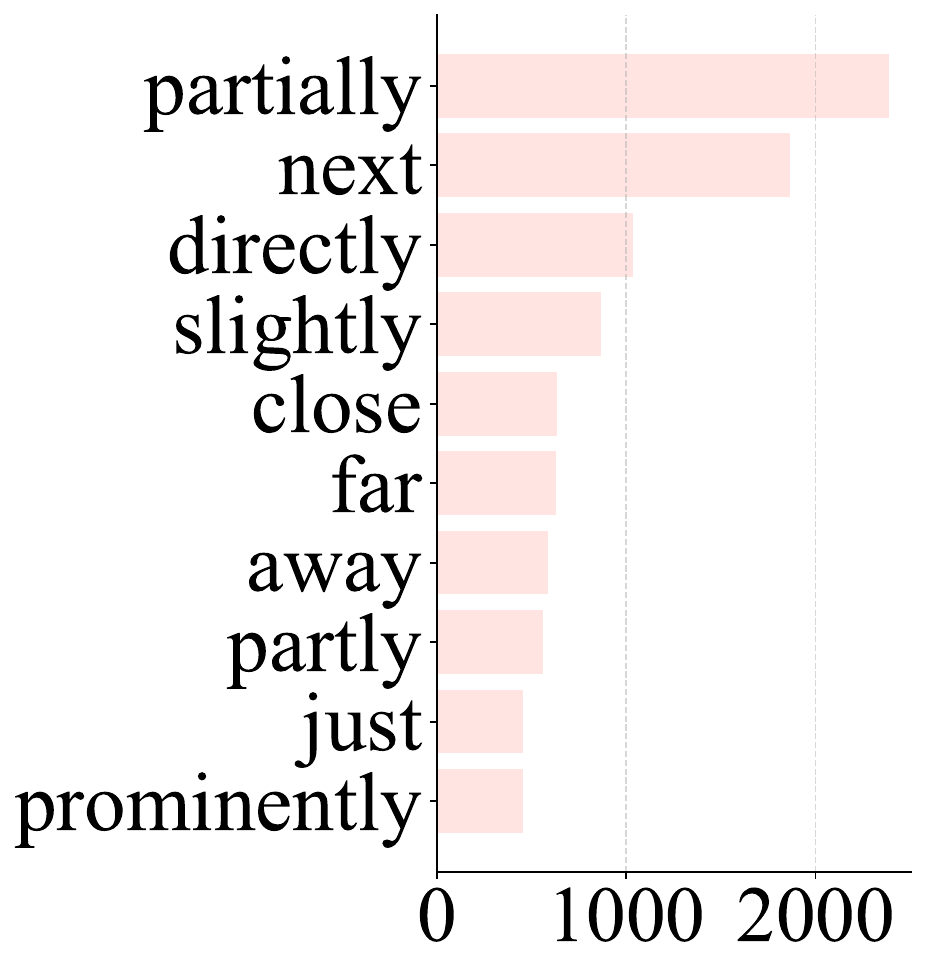}
        \caption{Adverbs.}
    \end{subfigure}
    \begin{subfigure}[b]{0.23\textwidth}
        \centering
        \includegraphics[width=\textwidth]{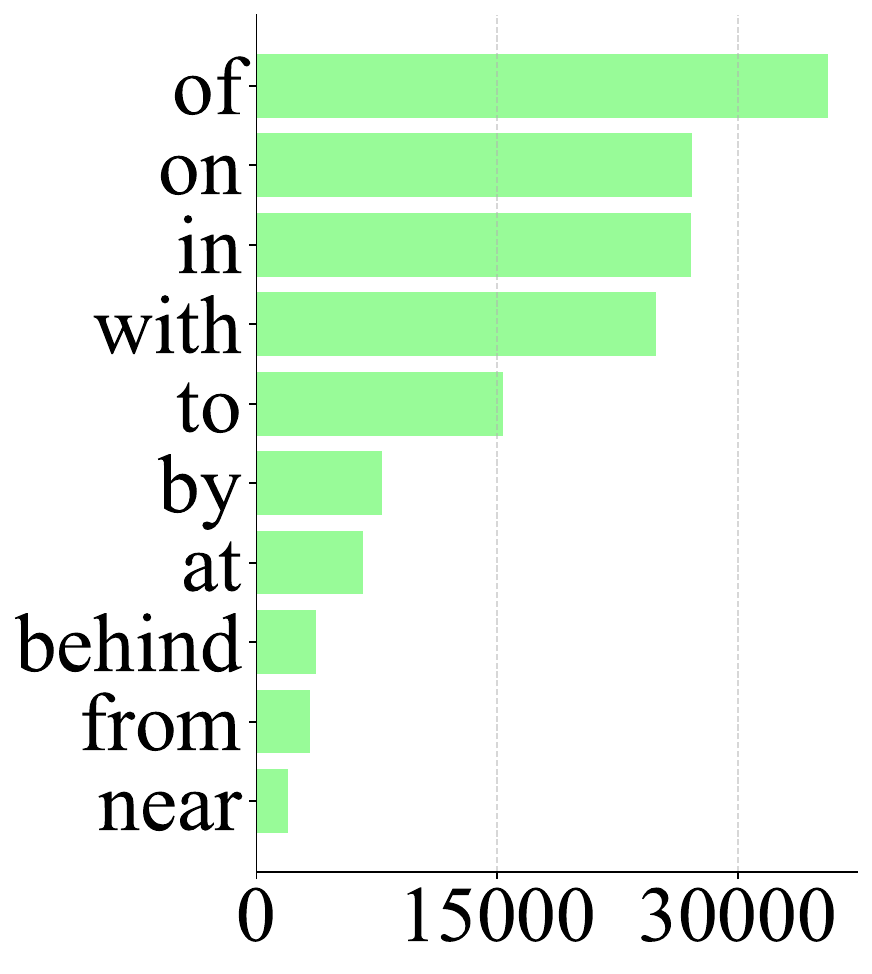}
        \caption{Prepositions.}
    \end{subfigure}
    \vspace{-3mm}
    \caption{The frequency of the 10 most frequently used words in each part-of-speech category, as parsed using the SpaCy library.}
    \label{fig:analysis-vocabulary}
\end{figure}

\textbf{Annotation Expansion.} To date, we have compiled 18,653 unique referring expressions, each describing a distinct instance. To assess the robustness of REC models to diverse language inputs, we employ a two-stage rephrasing process to expand our benchmark: 1) utilizing GPT-4 with the prompt detailed in Section~\ref{sec:prompt_rephrase}, to generate rephrased versions of each expression; 2) conducting a manual review to ensure that the rephrased expressions are unique, factual, relevant, and harmless. Consequently, our final Ref-L4 benchmark encompasses 9,735 images with 45,341 referring expressions, each accurately describing one of the 18,653 unique instances.

\subsection{Analysis}
\label{sec:analysis}

\textbf{Expression Length.} Figure~\ref{fig:exp_length_distri} illustrates the distribution of expression lengths across four different datasets: RefCOCO, RefCOCO+, RefCOCOg, and our Ref-L4. Due the the high overlap of data samples, RefCOCO and RefCOCO+ exhibit similar distributions, with a high density of shorter expressions peaking at around 3.6 words. RefCOCOg features slightly longer expressions on average, peaking at approximately 8.4 words. In contrast, our Ref-L4 displays a significantly different distribution, with expressions ranging much longer, peaking at around 24.2 words and having a long tail extending up to 117 words. This suggests that our Ref-L4 benchmark is designed to push the boundaries of current REC models, requiring them to process and comprehend more intricate and detailed descriptions.

\textbf{Instance Size.} In Figure~\ref{fig:box_size_distri}, we present a density plot comparing the instance sizes across four benchmarks. We define the instance size as the square root of the normalized size, $\sqrt{(hw)/(HW)}$, where $(h,w)$ represents the dimensions of the instance and $(H,W)$ represents the dimensions of the image. All benchmarks exhibit a peak density around an instance size of 160. Our Ref-L4 benchmark shows a wider distribution range compared to the other three, indicating that our Ref-L4 captures a broader spectrum of instance sizes.

\vspace{-1mm}
\textbf{Categories.} 
Our Ref-4L benchmark comprises 18,653 instances spanning 365 distinct categories, providing more complex and diverse evaluation scenarios. In contrast, RefCOCO and RefCOCO+ consists of 71 categories, while RefCOCOg covers 78 categories. Figure~\ref{fig:category_distribution} presents the distribution of instances among these 365 categories. Notably, the ten categories with the highest number of instances are ``Person'', ``Chair'', ``Hat'', ``Desk'', ``Lamp'', ``Cabinet/shelf'', ``Car'', ``Sneakers'', ``Handbag/Satchel'', and ```Flag''.

\vspace{-1mm}
\textbf{Vocabulary.} Our benchmark's referring expressions comprise a vocabulary totaling 22,813 unique words. This is significantly larger than the vocabulary sizes of RefCOCO, RefCOCO+, and RefCOCOg, which are 3,525, 4,387, and 5,050 words, respectively. Figure~\ref{fig:analysis-vocabulary} illustrates the 10 most frequently used nouns, verbs, adverbs, and prepositions.

\subsection{Evaluation}
\label{sec:evaluation}
\textbf{Evaluation Metrics.} We propose three distinct evaluation protocols:
\begin{enumerate}[leftmargin=*]
\item \textit{Accuracy.} This is the conventional metric used in REC. For a given referring expression and corresponding image, the target instance is considered successfully localized if the IoU between the predicted bounding box and the ground truth exceeds 0.5. Accuracy is then calculated as the ratio of successfully localized samples to the total number of samples, referred to as Acc$_{0.5}$ in this work. To better assess the localization capabilities of modern REC models, we also report accuracies at higher IoU thresholds: Acc$_{0.75}$, Acc$_{0.9}$, and mAcc, which is the average accuracy from Acc$_{0.5}$ to Acc$_{0.9}$ in increments of 0.05.
    \item \textit{Scale-Aware Performance.} To gain deeper insights into model capabilities, we report performance based on instance sizes: small, medium, and large. The size of an instance is defined as the square root of its area, $\sqrt{(hw)}$, where $(h,w)$ are the dimensions of the instance. Small instances are those with a size less than $128$, medium instances are between $128$ and $256$, and large instances exceed $256$. In total, there are $9 345$, $23 280$, and $12 716$ referring expressions describing $2,954$ small, $10,442$ medium, and $5,257$ large instances, respectively.
    \item \textit{Per-Category Performance.} Our benchmark encompasses a wide range of categories, up to $365$ in total. We provide an evaluation protocol to assess performance on a per-category basis.
\end{enumerate}

\textbf{Benchmark Division.} Modern large multimodal models (LMMs) that are able to handle the REC task typically use unrestricted and extensive data for training. Our Ref-L4 benchmark is designed to assess the capabilities of these advanced models without imposing any limitations on the training data sources. The benchmark is divided into two subsets: a validation set, comprising 30\% of the data with $7,231$ images, $10,311$ instances, and $13,420$ referring expressions; and a test set, comprising 70\% of the data with $9,467$ images, $17,242$ instances, and $31,921$ referring expressions. Given that our benchmark includes instances from $365$ categories, we ensure that each category has at least one sample in both the validation and test sets. While we provide these two splits, we encourage the combined use of both sets for model evaluation, especially in the current LMM era, where the use of unrestricted training data is prevalent.

\section{Experiments}
\label{src:exp}

\begin{table}[!t]
\centering
\small
\caption{Performance evaluation across 24 models on our Ref-L4 benchmark. NVIDIA A100 GPUs (80G) are utilized. The symbol $\dagger$ denotes models that outputs segmentation masks.}
\label{tab:perf_comparison}
\begin{tabular}{l|rrrrrr}
\toprule
\multirow{2.5}{*}{Model} &
  \multicolumn{4}{c}{Val+Test} &
  \multicolumn{1}{c}{Val} &
  \multicolumn{1}{c}{Test} \\ \cmidrule(l){2-5} \cmidrule(l){6-6} \cmidrule(l){7-7}
&
  Acc$_{0.5}$ &
  Acc$_{0.75}$ &
  Acc$_{0.9}$ &
  mAcc &
  mAcc &
  mAcc \\ \midrule
GPT-4V~\cite{gpt4,gpt4v,gpt4vcontribution}                 & 9.91  & 1.19  & 0.12  & 2.88  & 2.96  & 2.85  \\
KOSMOS-2~\cite{peng2023kosmos} & 48.53 & 38.34 & 17.54 & 34.72 & 34.89 & 34.64 \\
OFA-Tiny~\cite{wang2022ofa}          & 55.21 & 43.22 & 27.70 & 41.44 & 41.53 & 41.40 \\
OFA-Large~\cite{wang2022ofa}        & 72.53 & 62.31 & 45.02 & 59.17 & 59.42 & 59.07 \\
Ferret-7b~\cite{you2023ferret}      & 57.54 & 42.44 & 21.01 & 40.29 & 40.31 & 40.28 \\
Ferret-13b~\cite{you2023ferret}     & 64.44 & 49.04 & 27.46 & 46.88 & 47.31 & 46.71 \\
GroundingGPT~\cite{li2024lego}                & 60.84 & 40.48 & 12.00 & 38.19 & 38.42 & 38.09 \\
Shikra-7b~\cite{chen2023shikra}                   & 65.06 & 39.62 & 10.45 & 38.60 & 38.91 & 38.47 \\
Lenna~\cite{wei2023lenna}           & 65.90 & 58.55 & 45.58 & 55.69 & 55.88 & 55.60 \\
MiniGPTv2~\cite{chen2023minigpt}                   & 66.93 & 50.50 & 25.30 & 47.15 & 47.43 & 47.03 \\
Qwen-VL-Chat~\cite{bai2023qwen}                & 73.80 & 58.05 & 37.16 & 55.94 & 56.18 & 55.83 \\
ONE-PEACE~\cite{wang2023one}                   & 70.82 & 60.09 & 36.12 & 55.07 & 55.49 & 54.89 \\
SPHINX-MoE~\cite{gao2024sphinx}                  & 66.23 & 44.90 & 15.32 & 42.38 & 42.80 & 42.21 \\
SPHINX-MoE-1k~\cite{gao2024sphinx}               & 74.45 & 62.70 & 38.85 & 58.07 & 58.35 & 57.95 \\
SPHINX~\cite{lin2023sphinx}                      & 74.78 & 53.65 & 21.15 & 50.09 & 50.33 & 49.99 \\
SPHINX-1k~\cite{lin2023sphinx}                   & 78.52 & 62.17 & 32.95 & 57.57 & 57.91 & 57.42 \\
SPHINX-v2-1k~\cite{lin2023sphinx}                & 81.31 & 70.49 & 46.59 & 65.39 & 65.67 & 65.27 \\
CogVLM-Grounding~\cite{wang2023cogvlm}      & \textbf{81.70} & \textbf{70.77} & \textbf{48.35} & \textbf{66.09} & \textbf{66.25} & \textbf{66.02} \\ \midrule
PixelLM-7B$^\dagger$~\cite{ren2023pixellm}        & 41.83 & 27.57 & 13.32 & 27.10 & 27.09 & 27.11 \\
PixelLM-13B$^\dagger$~\cite{ren2023pixellm}       & 49.89 & 35.37 & 18.42 & 34.10 & 34.52 & 33.92 \\
LISA-Explanatory$^\dagger$~\cite{lai2023lisa}   & 65.12 & 52.35 & 38.26 & 50.77 & 50.89 & 50.72 \\
LISA$^\dagger$~\cite{lai2023lisa}              & 66.23 & 54.02 & 39.73 & 52.18 & 52.44 & 52.07 \\
PSALM$^\dagger$~\cite{zhang2024psalm}         & 67.26 & 58.22 & 44.11 & 55.46 & 55.68 & 55.37 \\
GlaMM$^\dagger$~\cite{rasheed2023glamm}        & \textbf{71.90} & \textbf{60.27} & \textbf{45.15} & \textbf{57.89} & \textbf{58.16} & \textbf{57.78} \\ \bottomrule
\end{tabular}
\end{table}

\textbf{Main Result.} We evaluate a total of 24 LMMs that can perform the REC task, dividing them into two categories based on their output type: those that produce bounding boxes and those that produce segmentation masks. For models that output segmentation masks, we convert these masks into tight bounding boxes to enable evaluation on our Ref-L4 benchmark. Table~\ref{tab:perf_comparison} presents the performance of these models on the validation set, test set, and the combined set, using the metrics defined in Section~\ref{sec:evaluation}. The evaluation prompt of GPT-4V is available in Section~\ref{sec:prompt_eval_gpt4-v}. Among the models that output bounding boxes, CogVLM-Grounding~\cite{wang2023cogvlm} shows the best performance, while GlaMM~\cite{rasheed2023glamm} leads in performance among the models that output masks.

\begin{figure}[!t]
    \centering
    \includegraphics[width=\textwidth]{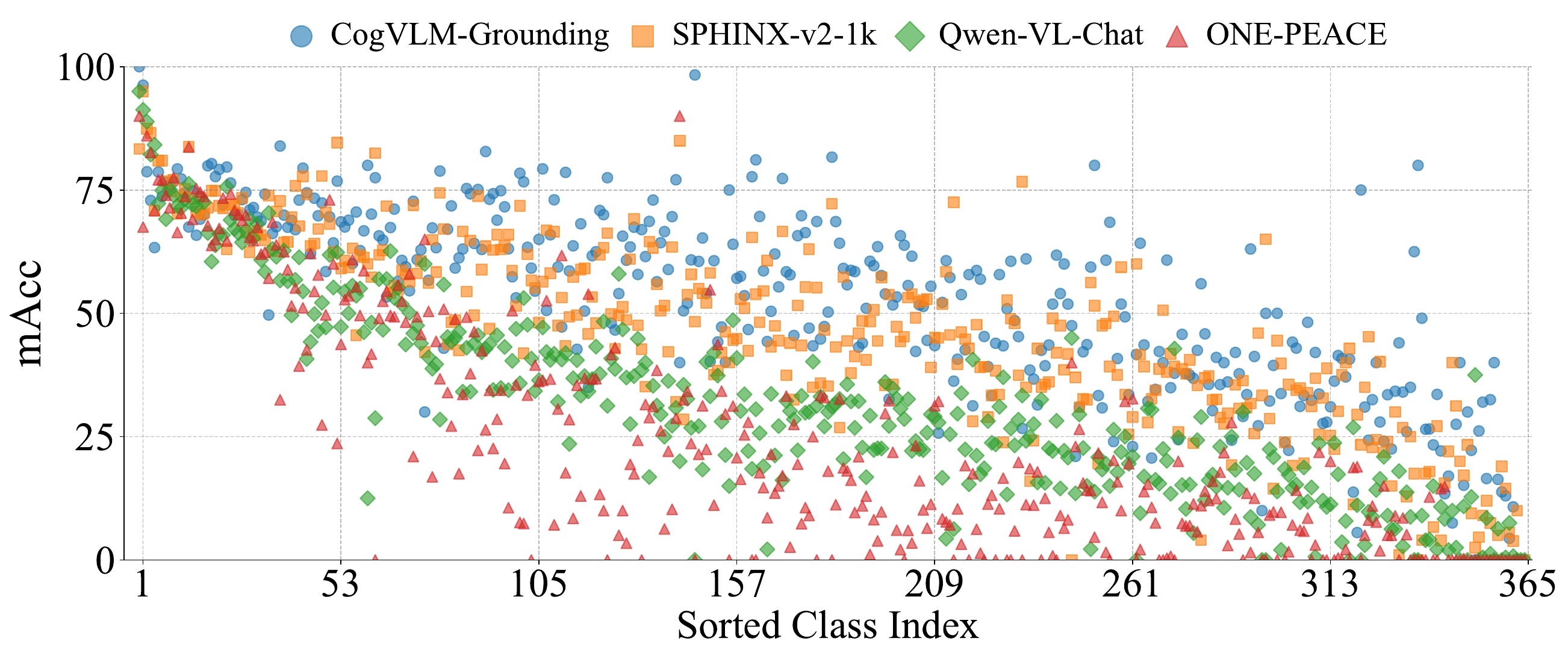}
    \vspace{-5mm}
    \caption{Category-wise performance of the four top-performing models on the val+test set, sorted in descending order based on their average per-category performance. The performance of all models can be found in Section~\ref{sec:appendix-category-wise}.}
    \label{fig:top4_models_mAcc_per_class}
\end{figure}

\begin{table}[!t]
\centering
\small
\caption{Scale-aware evaluation across 24 models on our Ref-L4 benchmark.}
\label{tab:perf_comparison_splits}
\begin{tabular}{l|rrrrrr}
\toprule
\multirow{2.5}{*}{Model} &
  \multicolumn{2}{c}{Small Size} &
  \multicolumn{2}{c}{Medium Size} &
  \multicolumn{2}{c}{Large Size} \\ 
  \cmidrule(l){2-3} \cmidrule(l){4-5} \cmidrule(l){6-7}
 &
  Acc$_{0.5}$ &
  mAcc &
  Acc$_{0.5}$ &
  mAcc &
  Acc$_{0.5}$ &
  mAcc \\ \midrule
GPT-4V~\cite{gpt4,gpt4v,gpt4vcontribution}              & 2.13    & 0.49    & 10.29   & 2.78    & 14.93   & 4.83    \\
KOSMOS-2~\cite{peng2023kosmos}            & 24.19   & 11.63   & 46.95   & 32.91   & 69.32   & 54.98   \\
OFA-Tiny~\cite{wang2022ofa}            & 17.91   & 11.49   & 65.13   & 49.00   & 64.46   & 49.61   \\
OFA-Large~\cite{wang2022ofa}           & 40.13   & 27.07   & 81.03   & 66.49   & 80.78   & 69.36   \\
Ferret-7b~\cite{you2023ferret}           & 30.93   & 14.57   & 62.40   & 43.72   & 68.18   & 52.92   \\
Ferret-13b~\cite{you2023ferret}          & 36.46   & 17.88   & 70.50   & 51.86   & 73.92   & 59.09   \\
GroundingGPT~\cite{li2024lego}        & 24.43   & 10.28   & 67.67   & 41.04   & 75.09   & 53.47   \\
Shikra-7b~\cite{chen2023shikra}           & 43.91   & 18.50   & 75.98   & 46.27   & 60.60   & 39.34   \\
Lenna~\cite{wei2023lenna}               & 31.02   & 23.48   & 72.90   & 61.53   & 78.72   & 68.66   \\
MiniGPTv2~\cite{chen2023minigpt}           & 32.99   & 14.85   & 73.67   & 51.16   & 79.52   & 63.53   \\
Qwen-VL-Chat~\cite{bai2023qwen}        & 47.66   & 26.26   & 79.80   & 61.06   & 82.01   & 68.37   \\
ONE-PEACE~\cite{wang2023one}           & 22.18   & 13.98   & 83.26   & 63.39   & 83.81   & 70.04   \\
SPHINX-MoE~\cite{gao2024sphinx}          & 39.48   & 16.39   & 72.97   & 46.38   & 73.55   & 54.17   \\
SPHINX-MoE-1k~\cite{gao2024sphinx}       & 58.96   & 37.61   & 77.80   & 61.53   & 79.70   & 66.77   \\
SPHINX~\cite{lin2023sphinx}              & 48.82   & 22.08   & 80.56   & 54.10   & 83.27   & 63.34   \\
SPHINX-1k~\cite{lin2023sphinx}           & 59.48   & 33.21   & 82.95   & 61.82   & 84.40   & 67.68   \\
SPHINX-v2-1k~\cite{lin2023sphinx}        & 65.23   & 43.43   & 84.00   & 68.45   & \textbf{88.21}   & \textbf{75.91}   \\
CogVLM-Grounding~\cite{wang2023cogvlm}    & \textbf{75.06}   & \textbf{52.85}   & \textbf{86.43}   & \textbf{71.31}   & 77.91   & 66.25   \\ \midrule
PixelLM-7B$^\dagger$~\cite{ren2023pixellm}          & 8.25    & 4.05    & 43.90   & 27.33   & 62.72   & 43.64   \\
PixelLM-13B$^\dagger$~\cite{ren2023pixellm}         & 17.05   & 8.54    & 53.40   & 35.48   & 67.59   & 50.34   \\
LISA-Explanatory$^\dagger$~\cite{lai2023lisa}    & 39.11   & 27.16   & 70.03   & 54.61   & 75.25   & 61.09   \\
LISA$^\dagger$~\cite{lai2023lisa}                & 39.24   & 27.49   & 71.17   & 56.05   & 77.01   & 63.22   \\
PSALM$^\dagger$~\cite{zhang2024psalm}               & 37.35   & 28.43   & 75.06   & 61.79   & 74.97   & 63.74   \\
GlaMM$^\dagger$~\cite{rasheed2023glamm}               & \textbf{47.07}   & \textbf{34.36}   & \textbf{77.17}   & \textbf{62.28}   & \textbf{80.50}   & \textbf{67.14}   \\ \bottomrule
\end{tabular}
\end{table}

\textbf{Category-Wise Performance.} Each instance in our benchmark is assigned a category label from one of 365 classes. Figure~\ref{fig:top4_models_mAcc_per_class} illustrates the performance of the top four models across these categories, sorted in descending order based on their average per-category performance. The results indicate a training bias issue, as all four models exhibit poor performance on some common categories.

\textbf{Scale-Aware Evaluation.} In Section~\ref{sec:evaluation}, we present a scale-aware evaluation to assess the model's ability to handle different instance scales. Specifically, we categorize all samples in our benchmark into three sets based on instance size: small, medium, and large. The performance of 24 models is detailed in Table~\ref{tab:perf_comparison_splits}. Among the bounding-box-output models, CogVLM-Grounding~\cite{wang2023cogvlm} excels with small and medium instances, while SPHINX-v2-1k~\cite{lin2023sphinx} achieves the best performance with large instances. For mask-output models, GlaMM~\cite{rasheed2023glamm} outperforms all other models across all three sets.

\begin{figure}[!t]
    \centering
    \includegraphics[width=.9\textwidth]{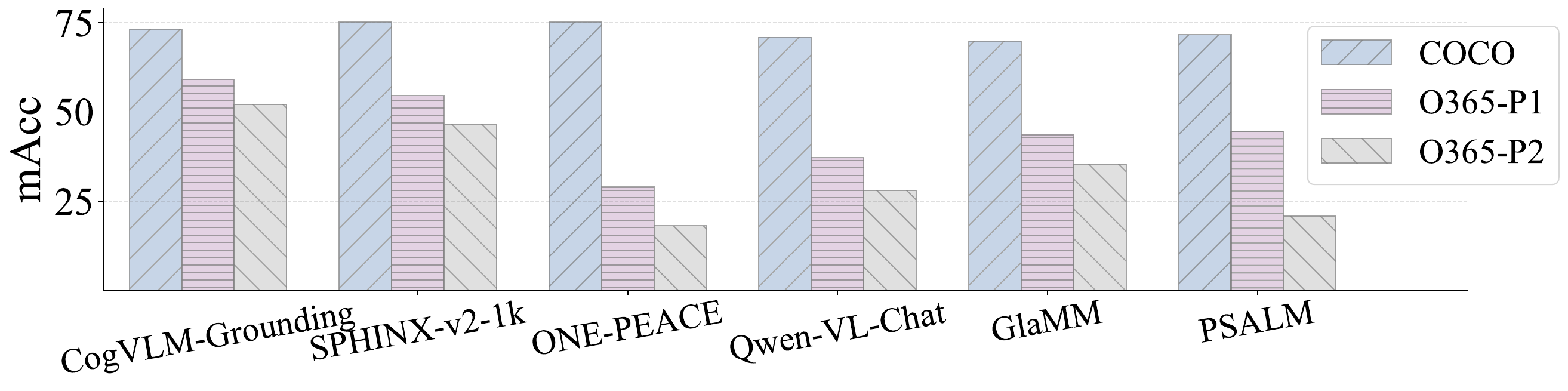}
    \caption{Evaluation of six models on various data sources, with mAcc acting as the metric. The results of all models can be found in Section~\ref{sec:appendix-data-sources}.}
    \label{fig:topX_models_mAcc}
\end{figure}

\textbf{Evaluation on Diverse Data Sources.} Our benchmark is derived from COCO and Objects365 datasets. We assess the performance of the top four models with bounding box outputs and the top two models with mask outputs across various subsets originating from either COCO or Objects365. These subsets are: 1) the COCO-derived set (referred to as ``COCO''); 2) a subset from Objects365, where the instances have categories that also exist in COCO (referred to as ``O365-P1''); 3) another subset from Objects365, where the instances have categories not found in COCO (referred to as ``O365-P2''). Figure~\ref{fig:topX_models_mAcc} presents the performance of these models across the three subsets. The ``COCO'' set shows higher accuracy compared to the other two sets, partially because most models are trained on the RefCOCO series and have limited exposure to Objects365 images. ``O365-P1'' exhibits higher accuracy than ``O365-P2'', as the latter includes more rare categories.

\section{Conclusion}
In this work, we first point out several limitations of the current REC benchmarks, such as substantial labeling inaccuracies and very brief referring expressions. To better assess the capabilities of models, particularly those LMMs that can perform the REC task, we present Ref-L4, which features four key characteristics: 1) a large-scale dataset with 45,341 annotations; 2) a wide range of object categories and varying instance scales; 3) detailed referring expressions; and 4) an extensive vocabulary comprising 22,813 unique words. We evaluate a total of 24 models using various evaluation protocols. We wish that Ref-L4 could serve as a valuable resource for researchers and developers, fostering the development of more robust and versatile REC models in the LMM era.

{\small
\bibliographystyle{abbrvnat}
\bibliography{ref}
}


\clearpage
\newpage
\appendix
\section{Labeling Errors in Existing Benchmarks}
\label{sec:label_error}
\begin{figure}[!h]
    \centering
    \newcommand{\SubFigWidth}{0.32}
    \begin{subfigure}[b]{\SubFigWidth\textwidth}
        \centering
        \includegraphics[width=\textwidth]{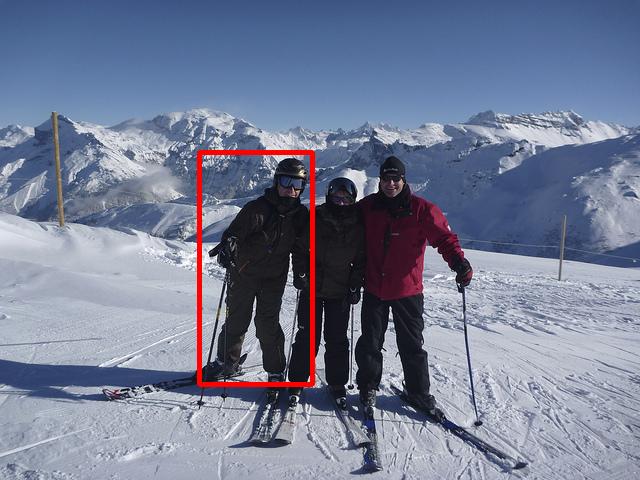}
        \caption{\input{refcoco_errors/type1/COCO_train2014_000000023539_457864_16264_refcoco+_testA.txt}}
    \end{subfigure}
    \begin{subfigure}[b]{\SubFigWidth\textwidth}
        \centering
        \includegraphics[width=\textwidth]{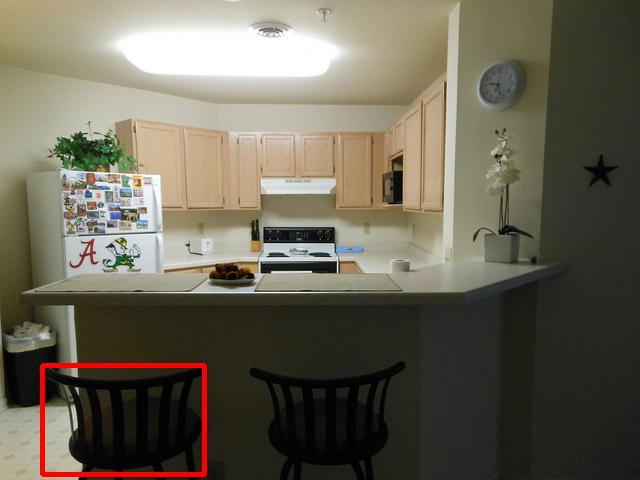}
        \caption{\input{refcoco_errors/type1/COCO_train2014_000000002448_385873_7632_refcocog_test.txt}}
    \end{subfigure}
    \begin{subfigure}[b]{\SubFigWidth\textwidth}
        \centering
        \includegraphics[width=\textwidth]{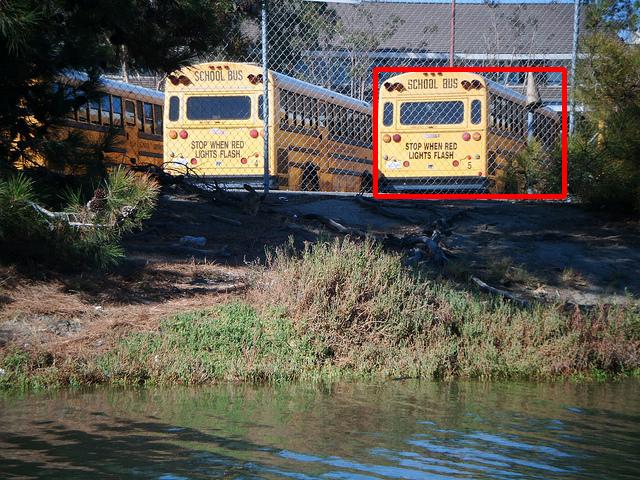}
        \caption{\input{refcoco_errors/type1/COCO_train2014_000000191754_164447_19763_refcoco+_testB.txt}}
    \end{subfigure}
    
    \begin{subfigure}[b]{\SubFigWidth\textwidth}
        \centering
        \includegraphics[width=\textwidth]{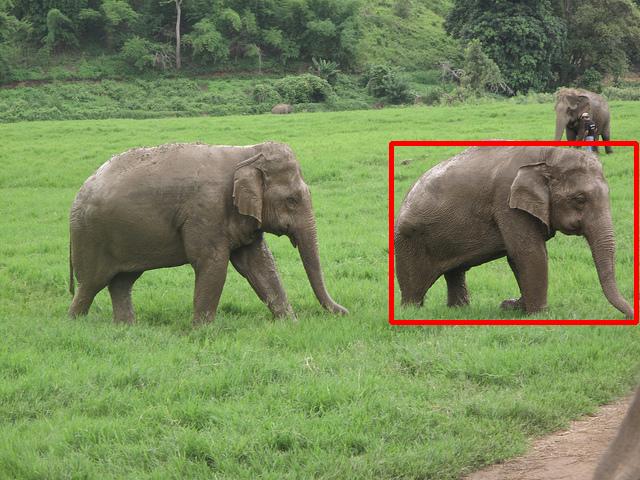}
        \caption{\input{refcoco_errors/type1/COCO_train2014_000000447934_580120_6969_refcocog_test.txt}}
    \end{subfigure}
    \begin{subfigure}[b]{\SubFigWidth\textwidth}
        \centering
        \includegraphics[width=\textwidth]{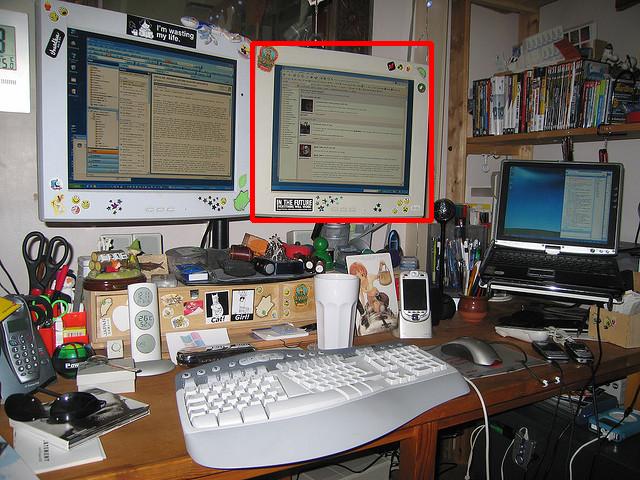}
        \caption{\input{refcoco_errors/type1/COCO_train2014_000000232167_2134255_3386_refcocog_test.txt}}
    \end{subfigure}
        \begin{subfigure}[b]{\SubFigWidth\textwidth}
        \centering
        \includegraphics[width=\textwidth]{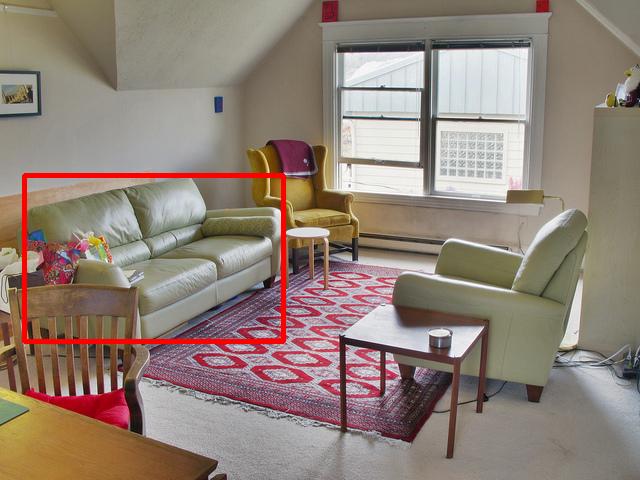}    \caption{\input{refcoco_errors/type3/COCO_train2014_000000015485_115263_21228_refcoco+_testB.txt}}
    \end{subfigure}
    \caption{Visualization of labeling errors, where a referring expression refers to multiple instances within the same image. For each sub-figure, we display the original bounding box annotation with a red rectangle and include the corresponding referring expression in the caption.}
    \label{fig:refcoco_error_type1}
\end{figure}

\begin{figure}[!h]
    \centering
    \newcommand{\SubFigWidth}{0.32}
    \begin{subfigure}[b]{\SubFigWidth\textwidth}
        \centering
        \includegraphics[width=\textwidth]{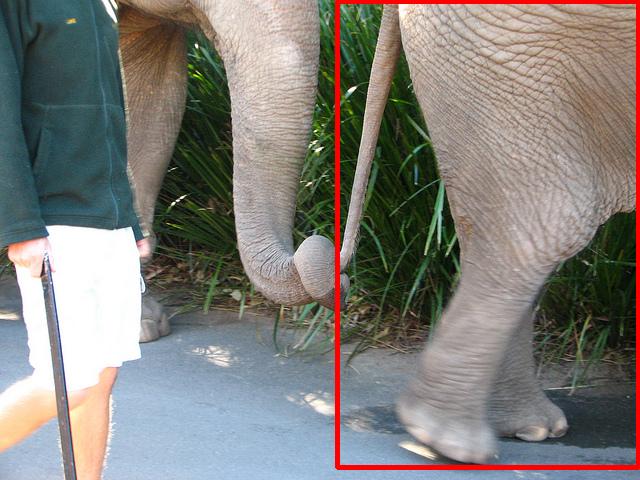}
        \caption{\input{refcoco_errors/type2/COCO_train2014_000000013763_583857_21243_refcoco+_testB.txt}}
    \end{subfigure}
    \begin{subfigure}[b]{\SubFigWidth\textwidth}
        \centering
        \includegraphics[width=\textwidth]{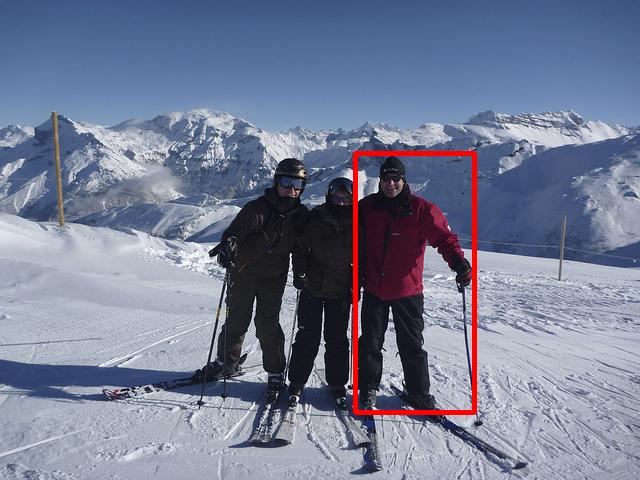}
        \caption{\input{refcoco_errors/type2/COCO_train2014_000000023539_440804_16267_refcoco+_testA.txt}}
    \end{subfigure}
    \begin{subfigure}[b]{\SubFigWidth\textwidth}
        \centering
        \includegraphics[width=\textwidth]{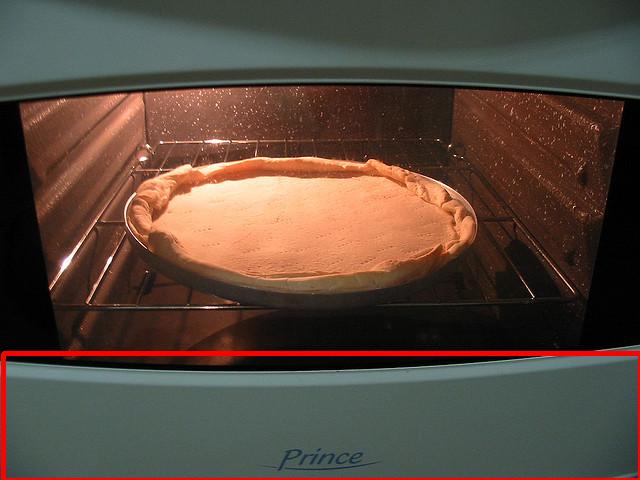}
        \caption{\input{refcoco_errors/type2/COCO_train2014_000000027679_1127893_21130_refcoco+_testB.txt}}
    \end{subfigure}
    
    \begin{subfigure}[b]{\SubFigWidth\textwidth}
        \centering
        \includegraphics[width=\textwidth]{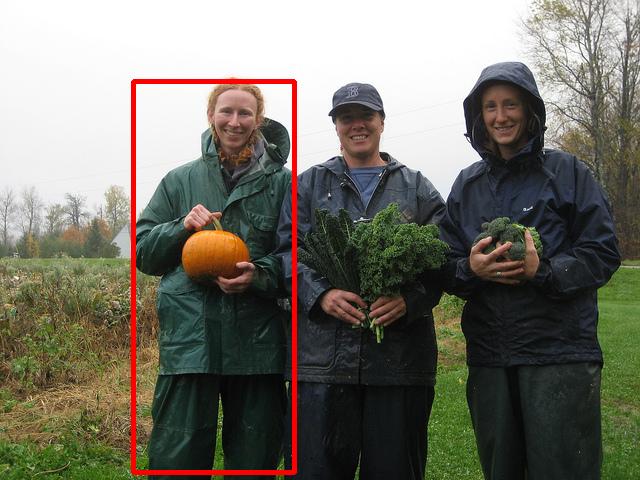}
        \caption{\input{refcoco_errors/type2/COCO_train2014_000000029799_542728_16184_refcoco+_testA.txt}}
    \end{subfigure}
    \begin{subfigure}[b]{\SubFigWidth\textwidth}
        \centering
        \includegraphics[width=\textwidth]{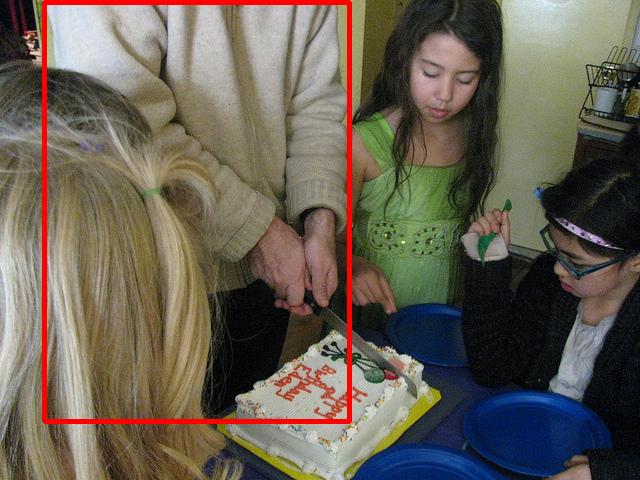}
        \caption{\input{refcoco_errors/type2/COCO_train2014_000000031187_1732500_347128_refcocog_val.txt}}
    \end{subfigure}
    \begin{subfigure}[b]{\SubFigWidth\textwidth}
        \centering
        \includegraphics[width=\textwidth]{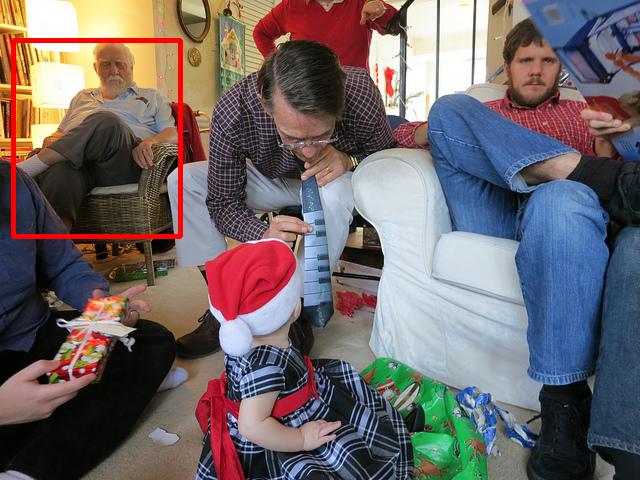}
        \caption{\input{refcoco_errors/type2/COCO_train2014_000000048150_508634_15965_refcoco+_testA.txt}}
    \end{subfigure}
    
    \caption{Visualization of labeling errors, where the bounding box annotations are inaccurate. For each sub-figure, we display the original bounding box annotation with a red rectangle and include the corresponding referring expression in the caption.}
    \label{fig:refcoco_error_type2}
\end{figure}

\begin{figure}[!h]
    \centering
    \newcommand{\SubFigWidth}{0.32}
    \begin{subfigure}[b]{\SubFigWidth\textwidth}
        \centering
        \includegraphics[width=\textwidth]{refcoco_errors/type1/COCO_train2014_000000427555_531968_1441_refcoco_testA.jpg}      \caption{\input{refcoco_errors/type1/COCO_train2014_000000427555_531968_1441_refcoco_testA.txt}}
    \end{subfigure}
    \begin{subfigure}[b]{\SubFigWidth\textwidth}
        \centering
        \includegraphics[width=\textwidth]{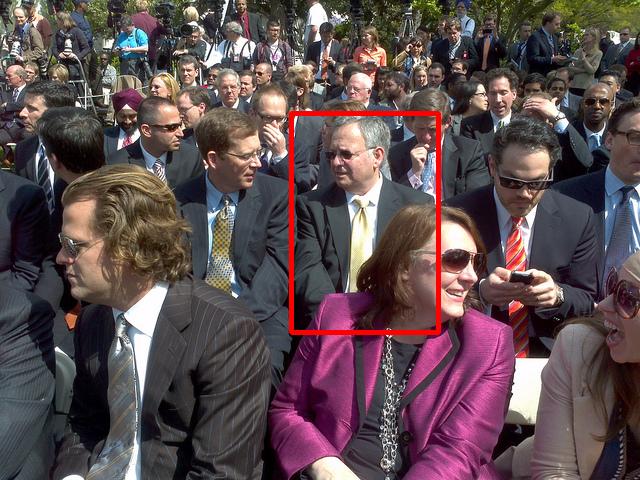}
        \caption{\input{refcoco_errors/type3/COCO_train2014_000000039195_1279369_5253_refcoco_testA.txt}}
    \end{subfigure}
    \begin{subfigure}[b]{\SubFigWidth\textwidth}
        \centering
        \includegraphics[width=\textwidth]{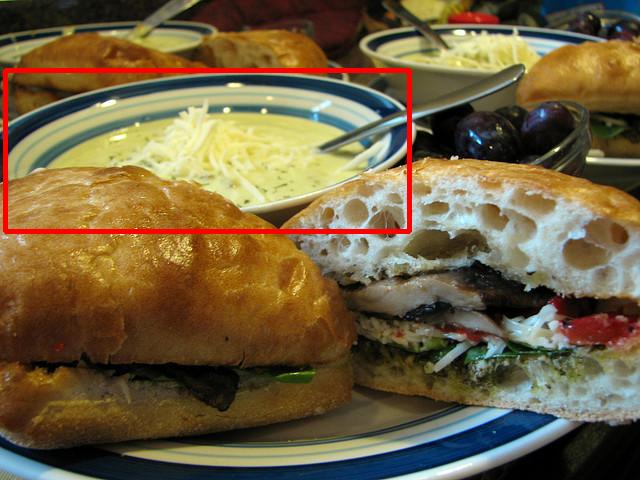}
        \caption{\input{refcoco_errors/type3/COCO_train2014_000000046997_1037915_10315_refcoco_testB.txt}}
    \end{subfigure}
    
    \begin{subfigure}[t]{\SubFigWidth\textwidth}
        \centering
        \includegraphics[width=\textwidth]{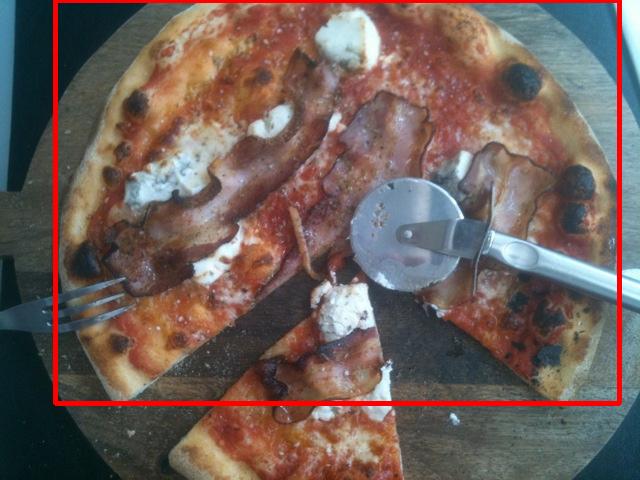}
        \caption{\input{refcoco_errors/type3/COCO_train2014_000000144320_1071868_20101_refcoco+_testB.txt}}
    \end{subfigure}
    \begin{subfigure}[t]{\SubFigWidth\textwidth}
        \centering
        \includegraphics[width=\textwidth]{refcoco_errors/type3/COCO_train2014_000000331331_254215_2607_refcoco_testA.jpg}
        \caption{\input{refcoco_errors/type3/COCO_train2014_000000331331_254215_2607_refcoco_testA.txt}}
    \end{subfigure}
    \begin{subfigure}[t]{\SubFigWidth\textwidth}
        \centering
        \includegraphics[width=\textwidth]{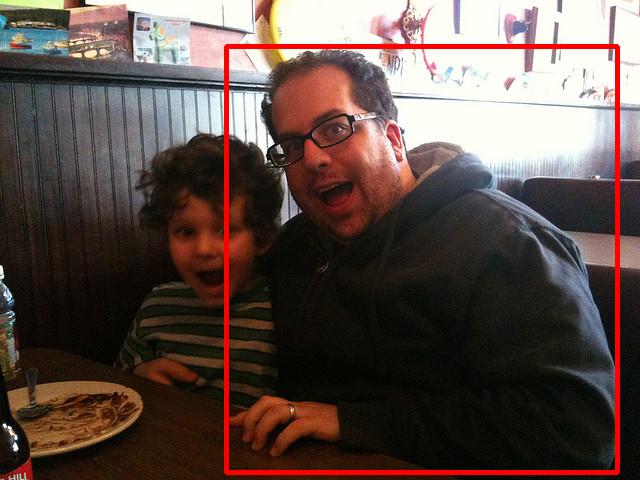}
        \caption{\input{refcoco_errors/type3/COCO_train2014_000000556698_459043_11012_refcoco+_testA.txt}}
    \end{subfigure}
    
    \caption{Visualization of labeling errors, where the referring expressions are either ambiguous or do not refer to any instance in the image. For each sub-figure, we display the original bounding box annotation with a red rectangle and include the corresponding referring expression in the caption.}
    \label{fig:refcoco_error_type3}
\end{figure}

In the REC task, a referring expression should uniquely describe an instance, which is represented by an accurate bounding box. We have identified and visualized three common types of labeling errors in the RefCOCO, RefCOCO+, and RefCOCOg benchmarks: 1) non-unique referring expressions (Figure~\ref{fig:refcoco_error_type1}), which refer to multiple instances within the same image; 2) inaccurate bounding boxes (Figure~\ref{fig:refcoco_error_type2}); and 3) misalignment between target instances and their referring expressions (Figure~\ref{fig:refcoco_error_type3}), where the referring expressions are either ambiguous or do not refer to any instance in the image.

\section{Prompts}

\subsection{Prompt for Context-Independent  Description Generation}
\label{sec:prompt_instance-level}

Briefly describe the [\textit{Category Name}] in one sentence. Begin your description with the object name, including adjectives if appropriate to describe its color or shape. Focus only on visible features and avoid mentioning blurriness.

Input image: [\textit{Cropped Image}].

\subsection{Prompt for Context-Aware Description Generation}
\label{sec:prompt_contextual}

You are a sophisticated referring expression generator. Your task is to generate a clear and specific description for the target instance highlighted by a red circle in the provided image, based on a given hint and the following criteria:

\textit{Criteria 1}: The description should enable individuals to understand and accurately identify the specified region within the image.

\textit{Criteria 2}: The description may should various attributes such as category, shape, size, color, visibility, exposure, texture, orientation, absolute position, relative position, facial features, clothing, accessories, gestures, context, semantic attributes, emotions, age, gender, posture, action, and especially interactions with other instances. The selection of features should be relevant to the particular region and the image context.

\textit{Criteria 3}: The red circle is solely for highlighting the region of interest. Do not refer to it in your descriptions.

\textit{Criteria 4}: Avoid using unnecessary words like ``look for'', ``spot'', ``observe'', ``find'', ``notice'', ``identify'', ``outline'', ``target'' and ``question''.

\textit{Criteria 5}: Ensure that the subject of each sentence matches the subject given in the hints. Do not incorrectly use the subject as the object.

\textit{Criteria 6}: Use the correct singular or plural form when referring to the target, which may be a single object, a pair of objects, or a group of objects.

\textit{Criteria 7}: Integrate all relevant information from the hints, noting that some hints may be redundant or contain errors.

Input image: [\textit{Raw Image}].

Hint: [\textit{Context-Independent Description}].

\subsection{Prompt for Rephrasing Referring Expressions}
\label{sec:prompt_rephrase}
Rewrite the subsequent description while preserving the main information. Utilize varied expressions and reorganize the sentences if necessary. Begin each sentence with the same subject being referred to.

Description: [\textit{The Referring Expression to be Rephrased}].

\subsection{Prompt for GPT4-V Evaluation}
\label{sec:prompt_eval_gpt4-v}

You are an expert in referring expression comprehension and localization. Your task is to locate the object in the image based on the provided expression. The coordinates range from the top left (0, 0) to the bottom right ([\textit{Image Width}], [\textit{Image Height}]). Please provide the bounding box in the format ($x_0$, $y_0$, $x_1$, $y_1$), where ($x_0$, $y_0$) represents the top-left corner and ($x_1$, $y_1$) represents the bottom-right corner.

Expression: [\textit{The Referring Expression}].

\section{More Experiments}

\begin{figure}[!t]
    \centering
    \begin{subfigure}[b]{\textwidth}
        \centering
        \includegraphics[width=0.95\textwidth]{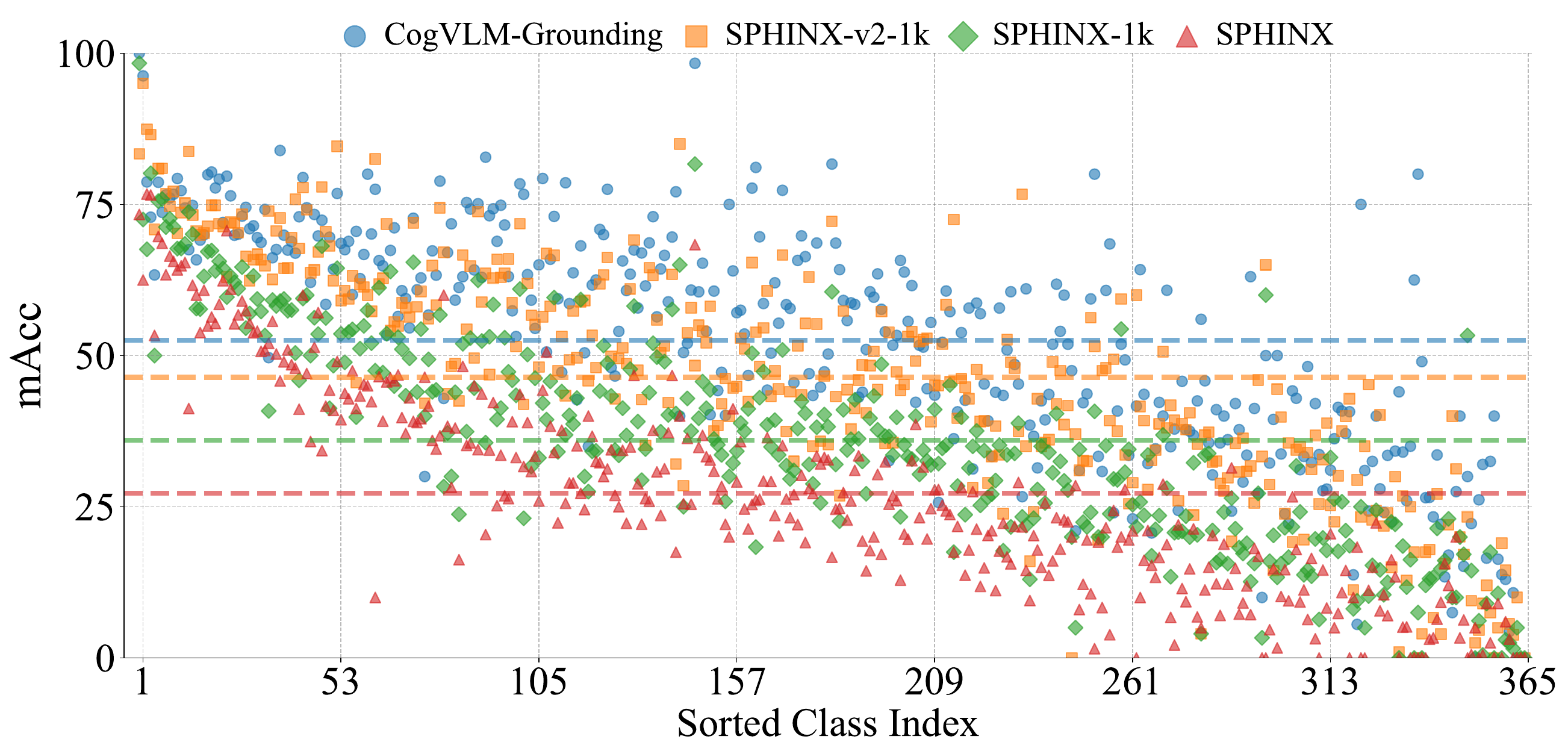}
        \caption{The average performance across all categories (dot lines) for CogVLM-Grounding~\cite{wang2023cogvlm}, SPHINX-v2-1k~\cite{lin2023sphinx}, SPHINX-1k~\cite{lin2023sphinx}, and SPHINX1~\cite{lin2023sphinx} are 52.56, 46.40, 36.01, and 26.95, respectively.}
    \end{subfigure}

    \begin{subfigure}[b]{\textwidth}
        \centering
        \includegraphics[width=0.95\textwidth]{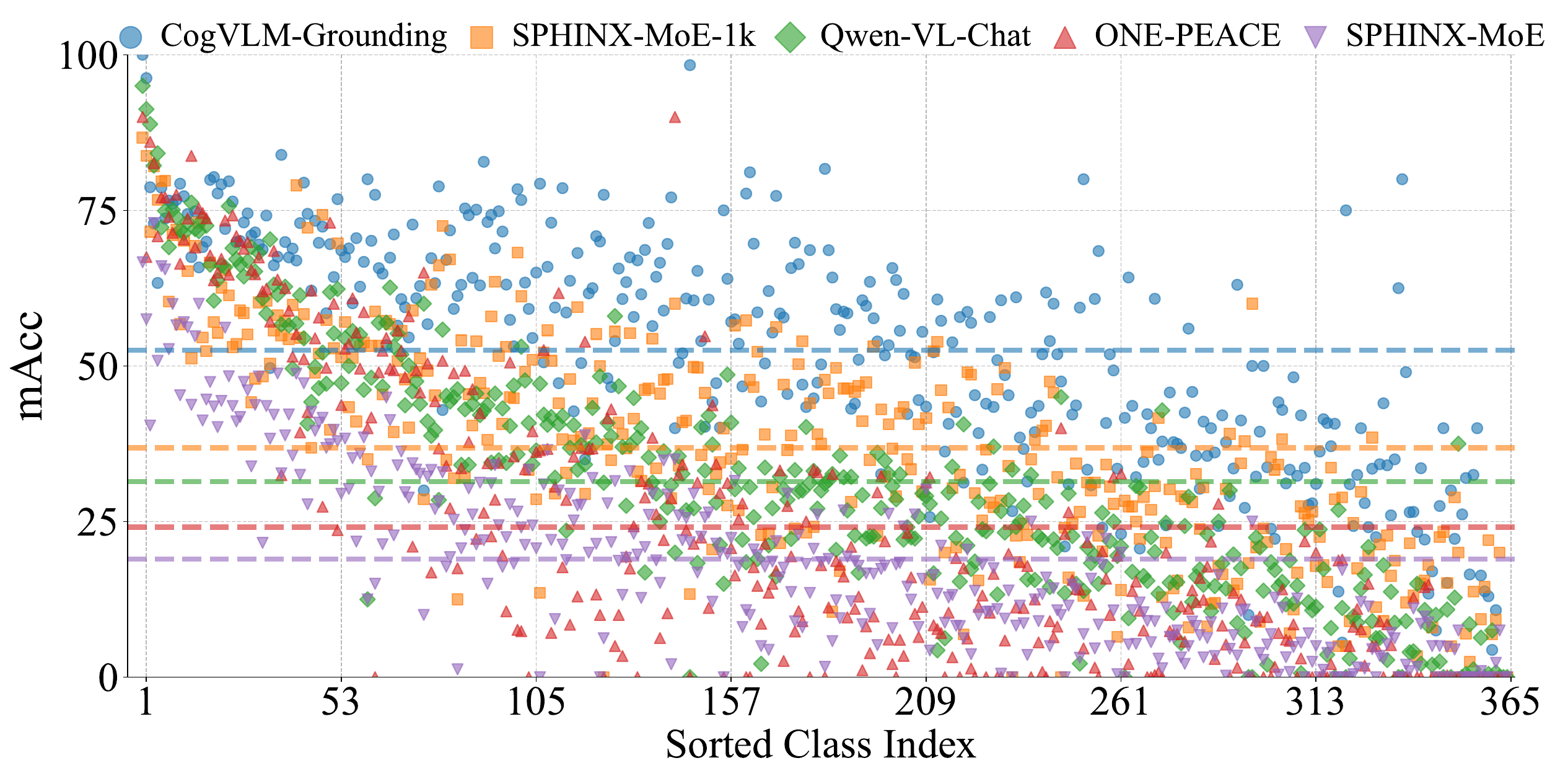}
        \caption{The average performance across all categories (dot lines) for SPHINX-MoE-1k~\cite{gao2024sphinx}, Qwen-VL-Chat~\cite{bai2023qwen}, ONE-PEACE~\cite{wang2023one}, and SPHINX-MoE~\cite{gao2024sphinx} are 36.84, 31.41, 24.11, and 18.77, respectively.}
    \end{subfigure}

    \begin{subfigure}[b]{\textwidth}
        \centering
        \includegraphics[width=0.95\textwidth]{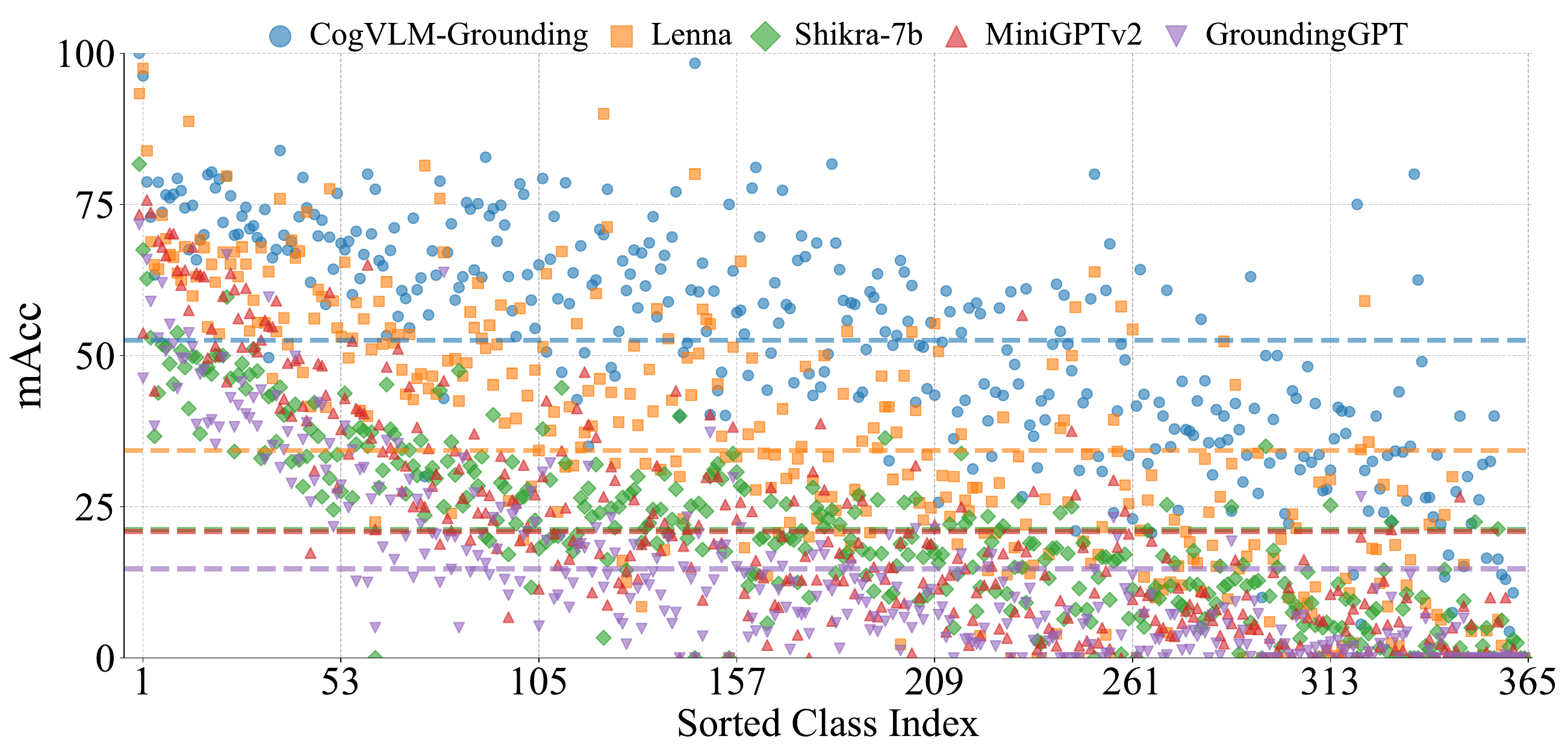}
        \caption{The average performance across all categories (dot lines) for Lenna~\cite{wei2023lenna}, Shikra-7b~\cite{chen2023shikra}, MiniGPTv2~\cite{chen2023minigpt}, and GroundingGPT~\cite{li2024lego} are 34.30, 21.22, 21.13, and 14.60, respectively.}
    \end{subfigure}
    
    \caption{Category-wise performance of 24 models (part-1), sorted in the same order as in Figure~\ref{fig:top4_models_mAcc_per_class}. We use CogVLM-Grounding as a reference for comparison in each sub-figure.}
    \label{fig:top24_models_mAcc_per_class_part1}
\end{figure}

\begin{figure}[!t]
    \centering
    \begin{subfigure}[b]{\textwidth}
        \centering
        \includegraphics[width=0.95\textwidth]{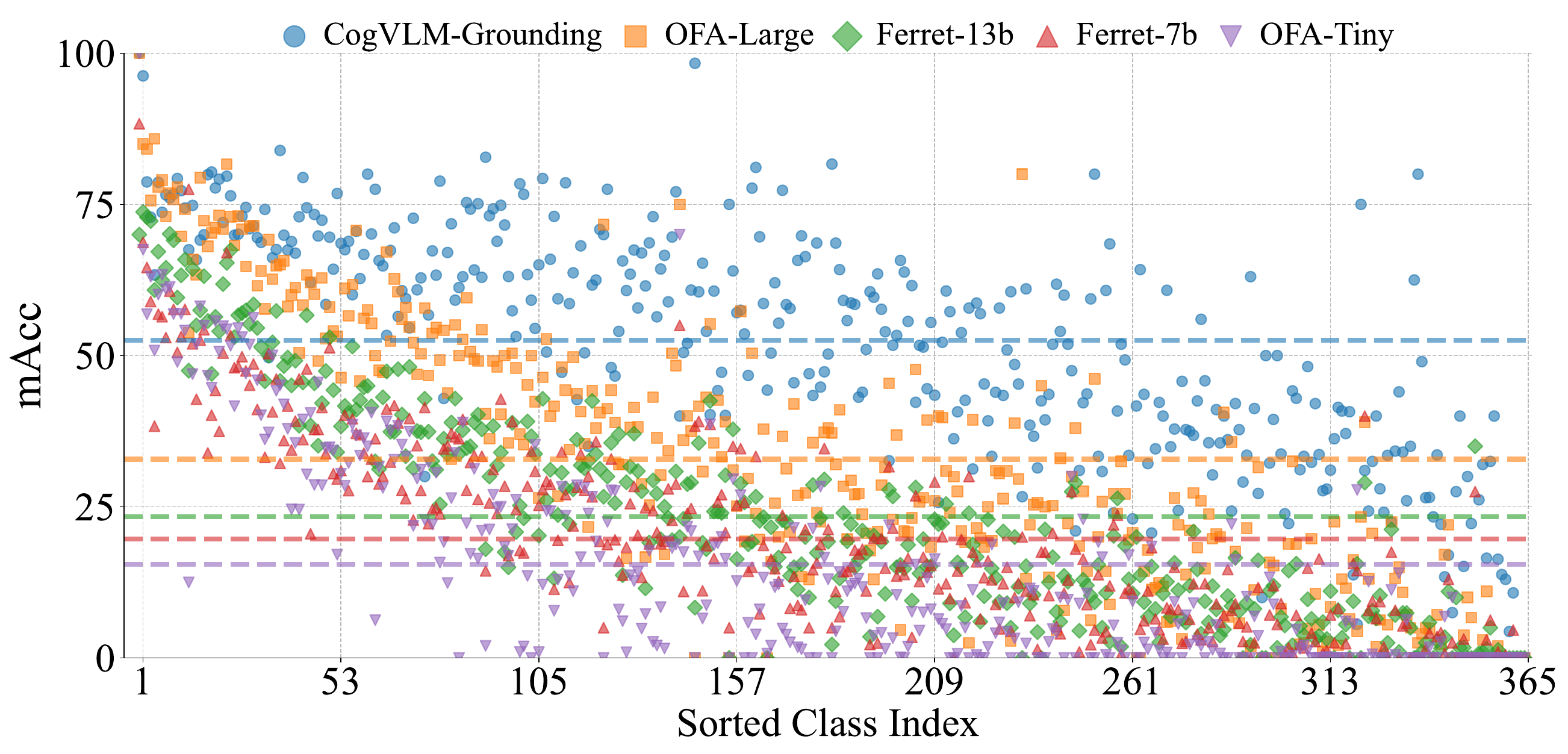}
        \caption{The average performance across all categories (dot lines) for OFA-Large~\cite{wang2022ofa}, Ferret-13b~\cite{you2023ferret}, Ferret-7b~\cite{you2023ferret} and OFA-Tiny~\cite{wang2022ofa} are 32.88, 23.33, 20.27, and 15.37, respectively.} 
    \end{subfigure}
    \begin{subfigure}[b]{\textwidth}
        \centering
        \includegraphics[width=0.95\textwidth]{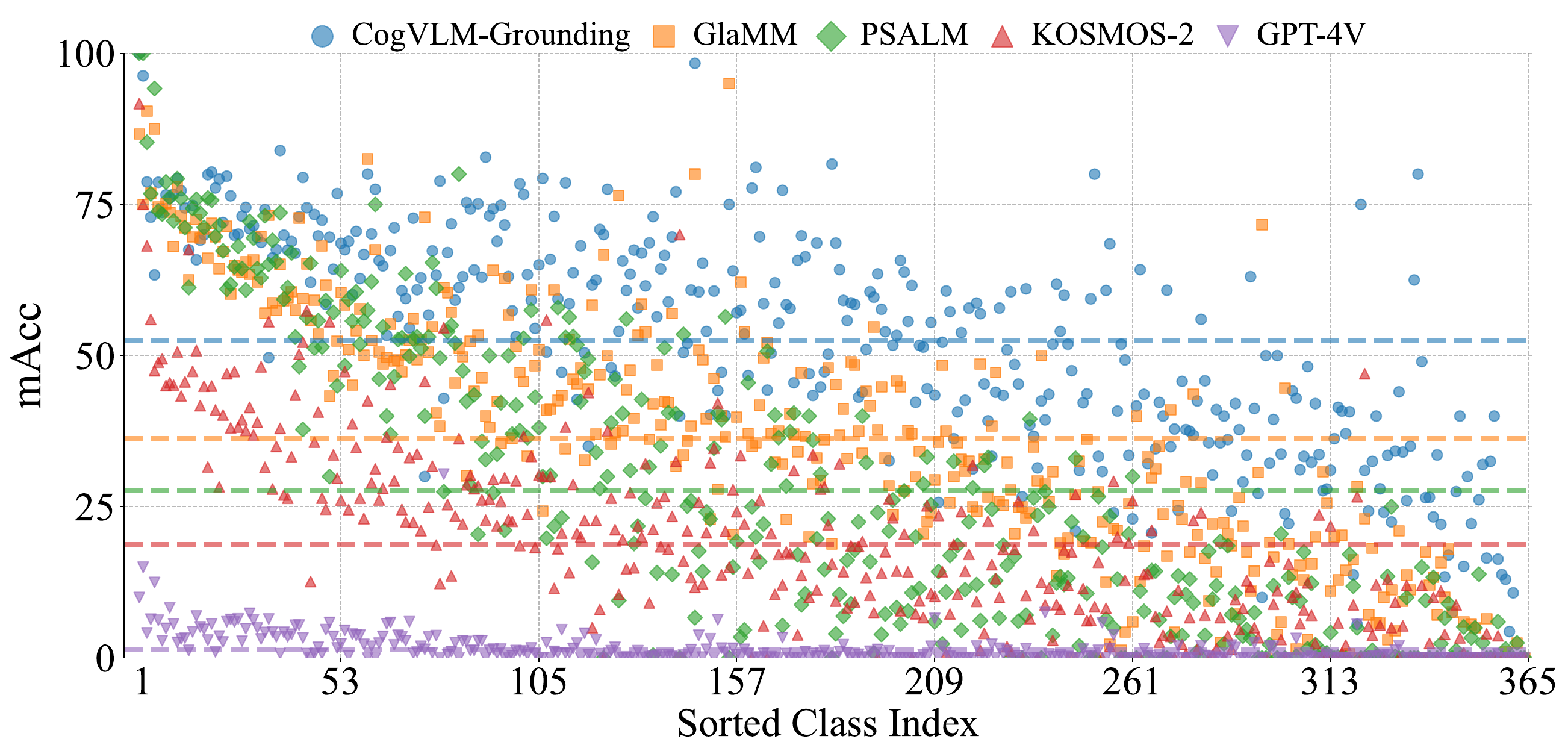}
        \caption{The average performance across all categories (dot lines) for GlaMM~\cite{rasheed2023glamm}, PSALM~\cite{zhang2024psalm}, KOSMOS-2~\cite{peng2023kosmos} and GPT-4V~\cite{gpt4,gpt4v,gpt4vcontribution} are 36.25, 27.62, 19.37, and 1.42, respectively.}
    \end{subfigure}
    \begin{subfigure}[b]{\textwidth}
        \centering
        \includegraphics[width=0.95\textwidth]{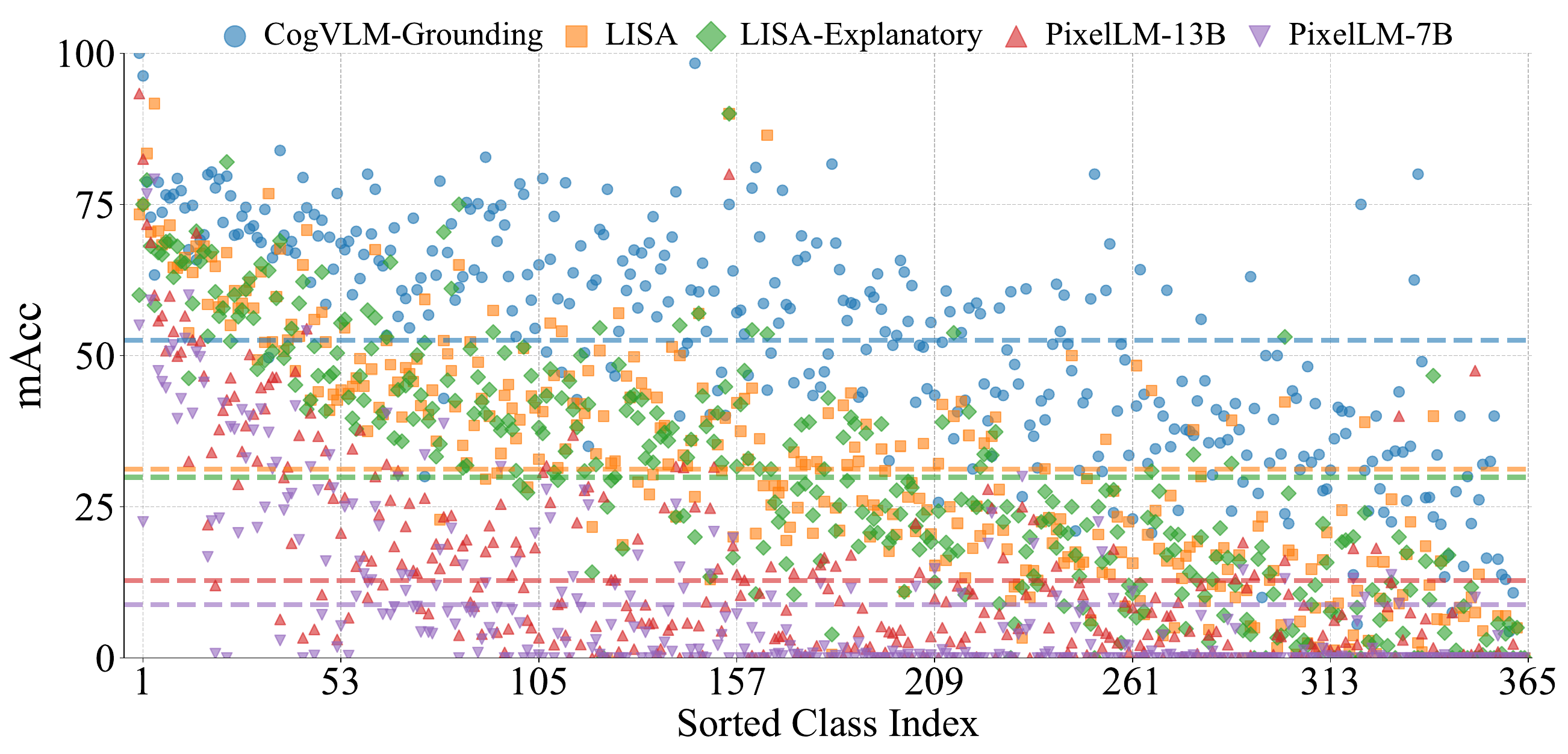}
        \caption{The average performance across all categories (dot lines) for LISA~\cite{lai2023lisa}, LISA-Explanatory~\cite{lai2023lisa}, PixelLM-13B~\cite{ren2023pixellm} and PixelLM-7B~\cite{ren2023pixellm} are 31.22, 29.87, 13.19, and 8.74, respectively.}
    \end{subfigure}
    \caption{Category-wise performance of 24 models (part-2), sorted in the same order as in Figure~\ref{fig:top4_models_mAcc_per_class}. We use CogVLM-Grounding as a reference for comparison in each sub-figure.}
    \label{fig:top24_models_mAcc_per_class_part2}
\end{figure}

\subsection{Category-Wise Performance.}
\label{sec:appendix-category-wise}
Figure~\ref{fig:top4_models_mAcc_per_class} presents the per-category performance of the top four models. In Figures~\ref{fig:top24_models_mAcc_per_class_part1} and \ref{fig:top24_models_mAcc_per_class_part2}, we show the performance for all 24 models on a per-category basis, with mAcc serving as the metric, along with the average performance for each model across all categories.

\begin{figure}[h]
    \centering
    \includegraphics[width=.95\textwidth]{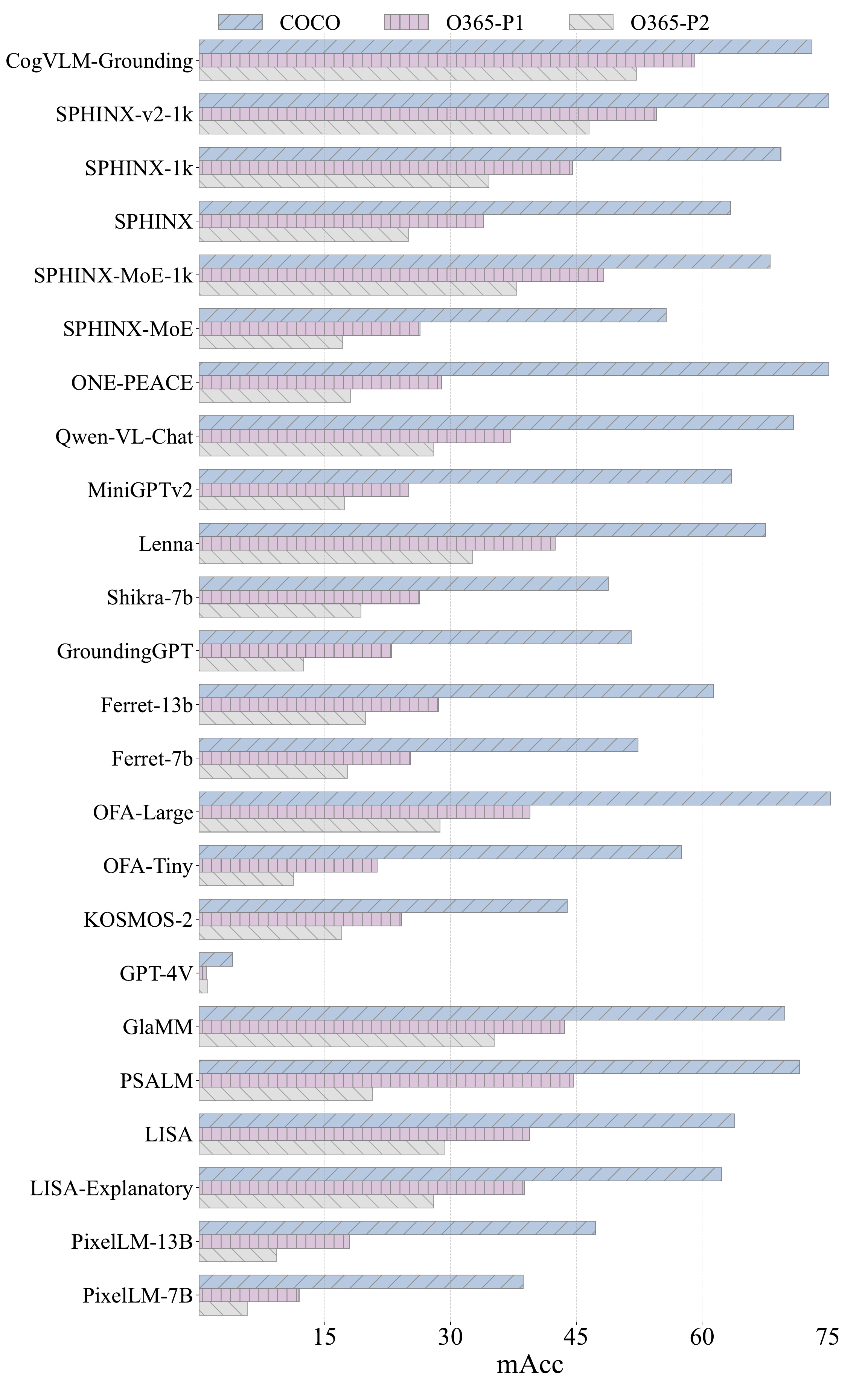}    
    \caption{Evaluation of 24 models on various data sources, with mAcc acting as the metric.}
    \label{fig:24-models_mAcc_data_source}
\end{figure}

\subsection{Evaluation on Diverse Data Sources.}
\label{sec:appendix-data-sources}
Figure~\ref{fig:topX_models_mAcc} illustrates the performance of six models across three subsets, namely ``COCO'', ``O365-P1'' and ``O365-P2''. In Figure~\ref{fig:24-models_mAcc_data_source}, the comprehensive results of 24 models across the same three subsets are displayed.

\section{Limitations and Broad Impacts}
\label{sec:limitations}
Ref-L4 provides a more comprehensive and detailed evaluation of REC capabilities, helping to better understand and improve the performance of large multimodal models capable of handling the REC task. The public availability of Ref-L4 and its evaluation code encourages further research and collaboration, driving innovation and advancements in the field of REC and beyond. While Ref-L4 aims to cover a wide range of scenarios, it may still miss out on specific edge cases or unique contexts that could be encountered in real-world applications. The detailed and lengthy referring expressions might pose a challenge for current models, requiring significant advancements in natural language processing and comprehension capabilities.


\end{document}